\documentclass[10pt,twocolumn,letterpaper]{article}

\usepackage{iccv}
\usepackage{cite}

\usepackage{enumerate}
\usepackage{times}
\usepackage{epsfig}
\usepackage{gensymb}
\usepackage{extarrows}
\usepackage{graphicx}
\usepackage{amsmath}
\usepackage{amssymb}
\usepackage{multirow}
\usepackage{tabularx}
\usepackage{tabularray}
\usepackage{booktabs}
\usepackage{ninecolors}
\usepackage{caption}
\usepackage{subcaption}

\usepackage{makecell}
\usepackage[export]{adjustbox}
\usepackage[normalem]{ulem}
\usepackage{textpos}  %

\usepackage{tabulary,overpic,xcolor,subfloat}

\newlength\savewidth
\newcommand{\dt}[1]{\fontsize{8pt}{0.1em}\selectfont (#1)}
\definecolor{ududff}{rgb}{0.30196078431372547,0.30196078431372547,1}

\definecolor{bgred}{rgb}{0.99, 0.89, 0.89}
\definecolor{bgblue}{rgb}{0.89, 0.89, 0.99}

\usepackage[pagebackref=true,breaklinks=true,letterpaper=true,colorlinks,bookmarks=false]{hyperref}

\iccvfinalcopy 

\def\iccvPaperID{****} 

\ificcvfinal\fi

\begin{document}

\title{DQS3D: Densely-matched Quantization-aware Semi-supervised 3D Detection}

\author{
Huan-ang Gao\textsuperscript{1,2} \quad
Beiwen Tian\textsuperscript{1,2} \quad
Pengfei Li\textsuperscript{1,2} \quad
Hao Zhao\textsuperscript{1} \quad
Guyue Zhou\textsuperscript{1} \\
\textsuperscript{1}Institute for AI Industry Research (AIR), THU \\
\textsuperscript{2}Department of Computer Science and Technology, THU \\
{\tt\small \{gha20, tbw18, li-pf22\}@mails.tsinghua.edu.cn} \quad {\tt\small \{zhaohao, zhouguyue\}@air.tsinghua.edu.cn}
}


\maketitle
\ificcvfinal\thispagestyle{empty}\fi

\begin{abstract}
   In this paper, we study the problem of semi-supervised 3D object detection, which is of great importance considering the high annotation cost for cluttered 3D indoor scenes. We resort to the robust and principled framework of self-teaching, which has triggered notable progress for semi-supervised learning recently. While this paradigm is natural for image-level or pixel-level prediction, adapting it to the detection problem is challenged by the issue of proposal matching. Prior methods are based upon two-stage pipelines, matching heuristically selected proposals generated in the first stage and resulting in spatially sparse training signals. In contrast, we propose the first semi-supervised 3D detection algorithm that works in the single-stage manner and allows spatially dense training signals. A fundamental issue of this new design is the quantization error caused by point-to-voxel discretization, which inevitably leads to misalignment between two transformed views in the voxel domain. To this end, we derive and implement closed-form rules that compensate this misalignment on-the-fly. Our results are significant, e.g., promoting ScanNet $\rm mAP@0.5$ from $35.2\%$ to $48.5\%$ using $20\%$ annotation. Codes and data are publicly available\footnote{Code: \textcolor{magenta}{\href{https://github.com/AIR-DISCOVER/DQS3D}{https://github.com/AIR-DISCOVER/DQS3D}}}.
\end{abstract}

\section{Introduction}

3D object detection (and reconstruction/tracking) \cite{song2014sliding, qi2019deep, chen2022pq, li2023rico, liu2022hoi4d, zhong2020seeing} is a fundamental problem in 3D scene understanding \cite{huang2018holistic, zou2019complete, li2022physically, zhao2017physics, zhao2021transferable, wu2022sc}, but its progress still lags behind 2D detection due to a high annotation cost. As such, semi-supervised 3D object detection \cite{zhao2020sess, wang20213dioumatch, yin2022semi} has recently attracted much attention as it holds the promise to improve accuracy using enormous unlabeled data. These semi-supervised 3D detectors are trained with a widely recognized framework called mean teachers (MT) \cite{tarvainen2017mean}. While semi-supervised image classification \cite{xie2020self} and semantic segmentation \cite{chen2021semi} using MT boil down to pairing predictions at the image or pixel level, how to pair predictions between two sets of 3D boxes remains an open question.

\begin{figure}[tbp]
    \centering
    \includegraphics[width=0.85\linewidth]{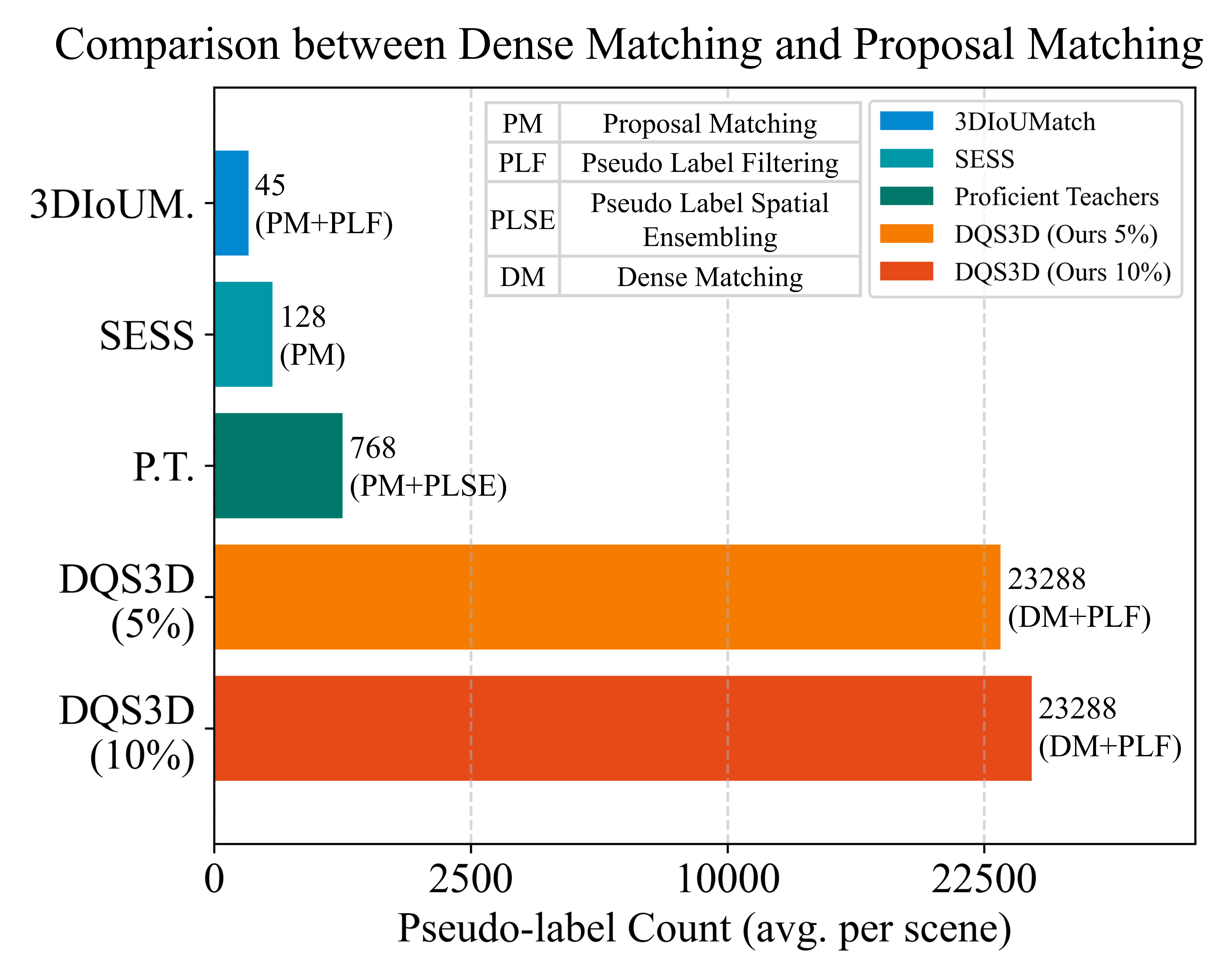}
    \caption{This figure demonstrates the average count of box pairs in representative proposal matching methods SESS \cite{zhao2020sess}, 3DIoUMatch \cite{wang20213dioumatch} and Proficient Teachers \cite{yin2022semi}. Our dense matching formulation (DQS3D) allows significantly more box pairs and spatially dense training signals. The x-axis is distorted according to a squared root mapping.}
    \label{fig:teaser}
    \vspace{-15pt}
\end{figure}

This open question is not yet well answered by prior methods \cite{zhao2020sess, wang20213dioumatch, yin2022semi}, as demonstrated by the analysis in Fig.~\ref{fig:teaser}. Shown by the upper three bars, they only exploit a very limited number of box pairs for MT training and we attribute this limitation to the two-stage architecture (i.e., VoteNet \cite{qi2019deep}) they are built upon. VoteNet makes final box predictions using seed proposals extracted by the first stage and only a limited number of proposal pairs are aligned.

\textbf{Being densely-matched.} The emergence of fully convolutional 3D detection \cite{rukhovich2022fcaf3d} inspires us to address the aforementioned issue using densely matched boxes, and it turns out fruitful. As shown by the lower two bars in Fig.~\ref{fig:teaser}, our method allows much more box pairs for MT training even after label filtering. This change leads to spatially dense training signals that translate to notable performance improvement (Table.~\ref{table:main}). In one word, our method predicts one 3D box for each voxel, getting rid of the intermediate proposal generation stage. Thus pairing teacher and student predictions in a voxel-wise manner becomes a natural choice and this directly leads to dense training signals.

\textbf{Being quantization-aware.} During the development of our densely matched paradigm, we identify a fundamental issue specific to 3D detection: point-to-voxel quantization. It is widely known that the power of MT is unleashed only with diverse data augmentation \cite{xie2020self, chen2021semi, gao2023semi} and random transformation is a typical augmentation strategy \cite{deng2021vector, huang2021spatio, tian2022vibus} for 3D point cloud. Unfortunately, applying random transformation inevitably leads to a different point-to-voxel mapping due to the existence of quantization error and a mismatch between teacher and student predictions on each voxel. To this end, we derive a closed-form compensation rule and implement it on-the-fly, which leads to consistent performance gains in various settings.

Highlighting our two technical contributions mentioned above, we name our method \textbf{DQS3D}, which is short for densely-matched quantization-aware semi-supervised 3D detection. Our contributions are summarized as follows:
\begin{itemize}
\item[$\bullet$] We shed light on the superiority of dense matching over proposal matching in semi-supervised 3D object detection, which could not only harvest more pseudo labels but also improve the pseudo-label quality.
\item[$\bullet$] We propose the first framework for densely-matched quantization-aware semi-supervised 3D object detection, where we point out the problem of quantization error and come up with an on-the-fly fix to it.
\item[$\bullet$] We conduct extensive experiments on public datasets and achieve significant results. For example, DQS3D scores $48.5\%$ $\rm mAP@0.5$ on ScanNet using $20\%$ data while the best published result is $35.2\%$. 
\end{itemize}

\section{Related Works}

\subsection{Self-Training for Semi-supervised Learning}
Semi-supervised learning (SSL) is a powerful learning paradigm that improves performance by leveraging both labeled and unlabeled data,
making it especially useful in situations where obtaining manually annotated labels is costly or difficult.
Recent works strive to apply this paradigm to tasks including semantic segmentation \cite{Kwon_2022_CVPR, Fan_2022_CVPR}, object detection \cite{Chen_2022_CVPR, Li_2022_CVPR, Mi_2022_CVPR}, text recognition \cite{Patel_2023_WACV}, action recognition \cite{Noguchi_2023_WACV}, facial expression recognition \cite{Li_2022_CVPR_facial}, video paragraph grounding \cite{Jiang_2022_CVPR}, \textit{etc}.

In particular, self-training using pseudo-labeling \cite{lee2013pseudo, zhao2020pointly} is a principled method that has been widely adopted for SSL \cite{Chu_2022_CVPR, Xiong_2021_ICCV, Feng_2021_CVPR, Wei_2021_CVPR, Yang_2021_CVPR, Duan_2021_CVPR, Zou_2021_ICCV, Gavrilyuk_2021_ICCV}.
A typical architecture for online self-training is mean teachers (MT) \cite{tarvainen2017mean}, which successfully integrates the self-training method into end-to-end frameworks. 
MT involves two identical but independent networks during training, with one (referred to as the student network) updated by gradient descent and the other one (referred to as the teacher network) updated by exponential moving average (EMA) of the student model's parameters.
Predictions of the teacher network on unlabeled data are regarded as online pseudo-labels for the student network, and self-teaching is implemented by enforcing predictions of the two networks to be consistent.
The architecture of MT has been proven highly effective on various tasks \cite{Chen_2022_CVPR, Liu_2022_CVPR, Li_2022_CVPR, Yang_2021_CVPR, Son_2021_ICCV, Uppal_2021_ICCV, Xu_2021_ICCV, Tang_2021_CVPR, Liu_2022_CVPR2}.

\subsection{Semi-supervised 3D Object Detection}

\textbf{Proposal Matching for Voting-based Detector.}
Specifically on the task of semi-supervised 3D object detection, numerous prior arts are also based on the MT architectures and take the voting-based VoteNet \cite{qi2019deep} as base detectors.
SESS \cite{zhao2020sess} introduced the nearest-center matching scheme (which we refer to as \textbf{\textit{proposal matching}}) to generate pseudo-labels from all teacher proposals. 
3DIoUMatch \cite{wang20213dioumatch} proposed a filtering mechanism to impose multiple thresholds on teacher predictions for improving quality of pseudo labels. It further performs non-maximum suppression (NMS) on pseudo-labels to reduce redundancy.
Proficient Teachers \cite{yin2022semi} implemented a spatial-ensembling module that generates detections from multiple augmented views of input point clouds, which are then combined to produce more pseudo-labels.
Although these methods have shown promise on the task of semi-supervised 3D object detection, they rely heavily on proposal matching, which we argue to be ineffective as the harvested pseudo training signals are sparse in space.

\textbf{Dense Prediction Detector.}
The dense prediction scheme for 2D object detection task has garnered a lot of interest in the research community \cite{tian2019fcos, law2018cornernet, redmon2016you, liu2016ssd}.
However, directly applying the backbones for 2D detection to 3D tasks \cite{liu2021group} is not cost-efficient due to the sparse nature of point clouds in space, requiring non-trivially larger amount of computational resources than 2D counterparts.

Nevertheless, the advent of high-dimensional convolutional neural networks \cite{choy20194d, choy2019fully, choy2020high, gwak2020gsdn, Zhao_2022_CVPR_code} has reduced both time and space complexity, making it possible to efficiently extract hierarchical features from 3D point clouds.
Leveraging sparse 3D convolution, 3D object detection can scale to much larger scenes while remaining memory-efficient \cite{liu2019point, wang2022cagroupd, rukhovich2022fcaf3d}.
Motivated by this design, FCAF3D \cite{rukhovich2022fcaf3d} uses a voxelized modification of ResNet as the backbone, which enables feature extraction and object prediction on a voxel basis.
The voxelization, however, inevitably brings about the issue of quantization error in the point-to-voxel discretization when the input point cloud is randomly augmented.
In this paper, we propose \textbf{\textit{dense matching} for dense prediction detector}, identify the problem of quantization error and propose a solution to it on-the-fly.


\section{Methodology}
This section presents a detailed exposition of DQS3D.
In Sec.~\ref{preliminary}, we formally define the task of semi-supervised 3D object detection.
In Sec.~\ref{dense}, we introduce dense matching scheme and compare it with prior arts of proposal matching.
In Sec.~\ref{framework}, we introduce the densely matched self-training framework and the loss design, combined to address the task of semi-supervised 3D object detection.
In Sec.~\ref{quantization}, we point out the problem of quantization error and derive a closed-form solution.

\subsection{Preliminary}
\label{preliminary}

We formally define the task of 3D object detection as to predict all objects $\mathbf{Y} = \{\mathbf{y}_i\}_{i=1}^K$ given an input point cloud $\mathbf{X} \in \mathbb{R}^{N \times 3}$, where $K$ denotes the number of objects in the scene and each target object $\mathbf{y}_i$ is represented by its bounding box parameters $\mathbf{\boldsymbol\delta}_i$ and corresponding semantic label $q_i$.
Specifically, in terms of 3d object detection in the semi-supervised setting, only a small proportion of the training dataset (denoted by $\{\mathbf{X}^L\}$) is equipped with ground-truth object bounding-box labels (denoted by $\{\mathbf{Y}^L\}$), whereas the remainder (denoted by $\{\mathbf{X}^U\}$) has no labels.


\begin{figure}[tbp]
    \centering
    \includegraphics[width=0.80\linewidth]{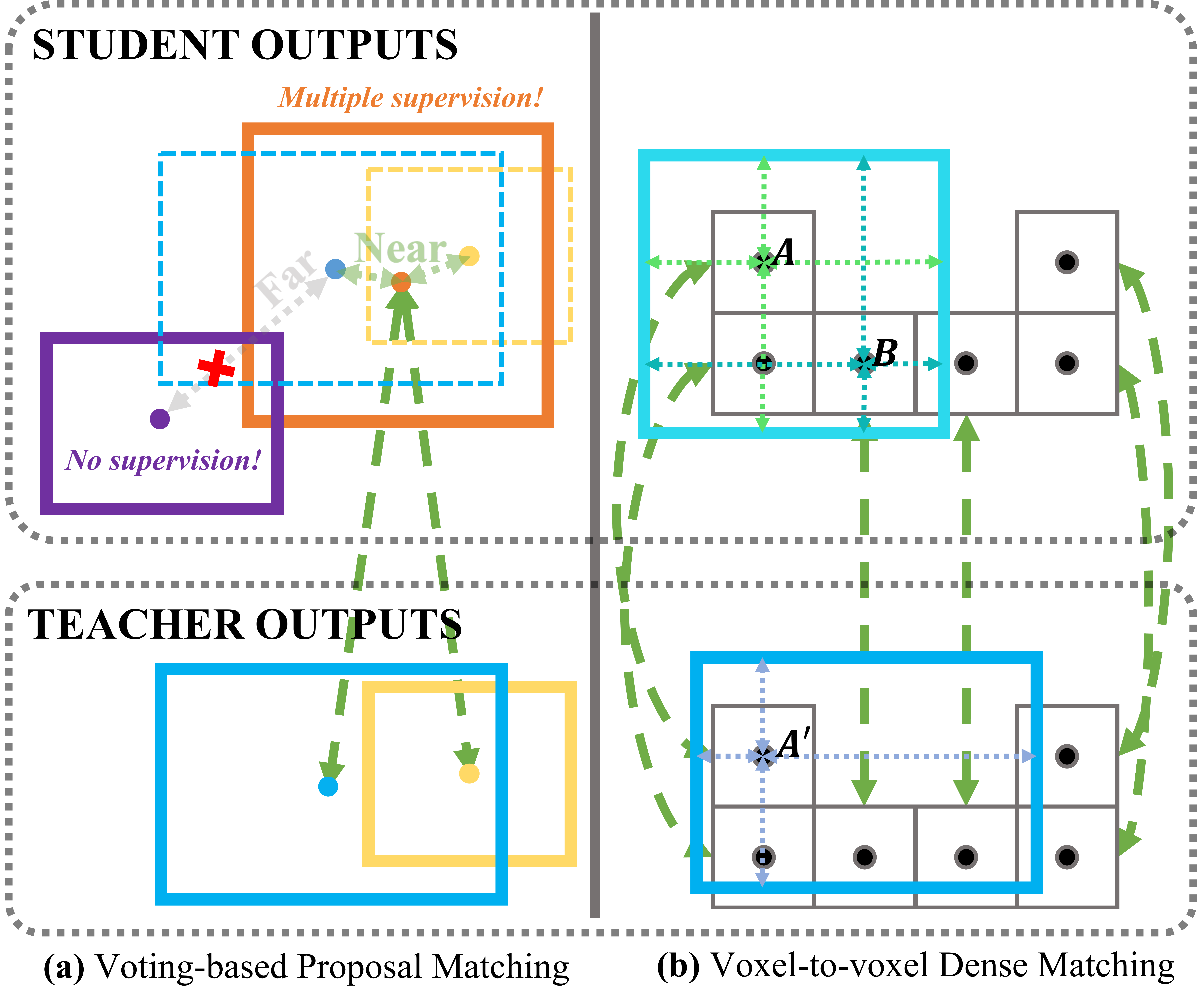}
    \caption{Illustration of two matching schemes. (a) Proposal matching: each teacher prediction is matched with the student prediction whose center is closest to that of the teacher prediction. (b) Dense matching: matching is established through spatially-aligned voxel anchors. Dashed boxes are only demonstrations for spatial locations of the teacher predictions.}
    \label{fig:dense-matching}
    \vspace{-10pt}
\end{figure}

\subsection{Dense Matching}
\label{dense}

To address the task of semi-supervised 3D object detection, self-training methods (e.g., mean teachers \cite{tarvainen2017mean, zhao2020sess, wang20213dioumatch, yin2022semi}) enforce consistency between the predictions of the student and teacher networks. Thus, it is crucial to establish a mapping between student and teacher predictions in aligned views, which we refer to as \textit{matching}.


Prior arts with self-training adopt \textit{proposal matching} \cite{zhao2020sess, wang20213dioumatch} to align the predicted objects (referred to as \textit{proposals}) of the student and teacher networks, which is typically done through a nearest-center strategy. More specifically, each teacher proposal is aligned with the student proposal whose center is the nearest to that of the teacher proposal, as illustrated in Fig.~\ref{fig:dense-matching}(a). 
Note that, the dashed boxes are only demonstration for spatial locations of teacher outputs. 

Despite the fact that teacher proposals are generally more accurate than student proposals, we argue that \textit{proposal matching} is ineffective and may hinder knowledge propagation from the teacher to the student. 
The ineffectiveness is mainly attributed to the two adverse situations illustrated in Fig.~\ref{fig:dense-matching}(a), inevitably caused by the sparseness of the proposals in space: \textcolor[rgb]{0.9257, 0.5, 0.1914}{\textbf{(1)} Adjacent teacher proposals are aligned to the same student proposal and cause confused supervision for the student.} \textcolor[rgb]{0.4375 0.1875 0.625}{\textbf{(2)} Student proposals that are distant from any teacher proposal are aligned to none and receive no supervision from teacher proposals}.

To address the aforementioned issues, a sufficient condition is a bijection between the student and teacher predictions.
Inspired by the facts that the objects are predicted on a voxel basis with the dense prediction base detector, and that the voxels anchors of the student and teacher views are inherently corresponded in space, we propose \textit{dense matching} (illustrated in Fig.~\ref{fig:dense-matching}(b)) to establish the bijection, simply by pairing the predictions at corresponding voxel anchors.

We believe that the dense matching scheme has the following advantages:
\textbf{(1)} 
Each predicted object is represented by multiple bounding box predictions whose regression scores varies at different voxel anchors (e.g., points $\mathbf A$ and $\mathbf B$ in Fig.~\ref{fig:dense-matching}(b)). This phenomenon imposes spatial regularization on the dense prediction model and improves the models's awareness of local geometry, as the optimization process forces the predicted bounding box parameters of the same object but at different voxel anchors to be sampled from a smooth function in space.
\textbf{(2)}
The required bijection between the student and teacher predictions is naturally established with the correspondence of the voxel anchors, with which each student prediction receives supervision from only one teacher prediction. This eliminates the two aforementioned adverse situations, namely the "\textcolor[rgb]{0.9257, 0.5, 0.1914}{multiple supervision}" and the "\textcolor[rgb]{0.4375 0.1875 0.625}{no supervision}".
In light of the benefits of dense matching over proposal matching, we propose a self-training framework specifically tailored for the dense matching scheme in the upcoming section.


\subsection{Densely Matched Self-Training Framework}

\label{framework}

\begin{figure}[tbp]
    \centering
    \includegraphics[width=0.85\linewidth]{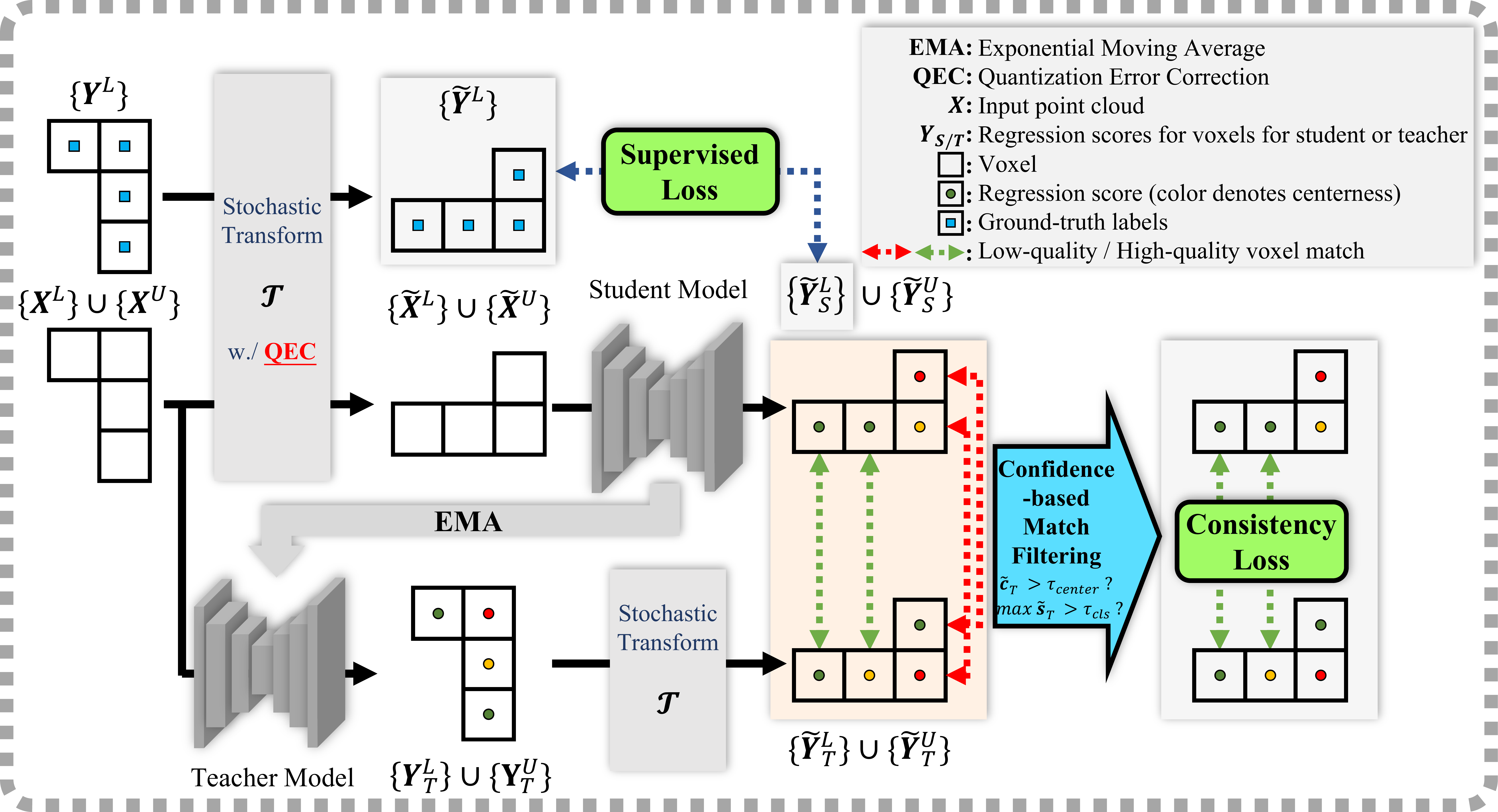}
    \caption{Illustration of our proposed densely-matched quantization-aware semi-supervised 3D object detection framework.}
    \label{fig:framework}
    \vspace{-15pt}
\end{figure}

\textbf{Overall Architecture.}
Following prior arts \cite{zhao2020sess, wang20213dioumatch, yin2022semi}, we adapt the robust self-training approach for our densely-matched semi-supervised learning framework, as depicted in Fig.~\ref{fig:framework}.
The framework includes two identical but independent networks (teacher and student models) as the base detectors, which are implemented by FCAF3D \cite{rukhovich2022fcaf3d}.
During training, an input batch is composed of both labeled data with ground-truth object annotations and unlabeled data.
The input batch is then augmented by asymmetric quantization-aware transformation $\mathbf{R}_{\theta, \Delta \mathbf{r}}$, which consists of a random rotation $\theta$ around the upright axis, a random translation $\Delta \mathbf{r}$ and the quantization error correction (detailed later in Sec.~\ref{quantization}).

The augmented and unaugmented batches are then fed into the student and teacher models, producing voxel-level predictions $\tilde{\mathbf{Y}}_S = \{\tilde{\mathbf{Y}}_S^L, \tilde{\mathbf{Y}}_S^U \}$ and ${\mathbf{Y}}_T = \{{\mathbf{Y}}_T^L, {\mathbf{Y}}_T^U \}$, respectively.
The teacher predictions are then aligned to the student predictions with the same transformation $\mathbf{R}_{\theta, \Delta \mathbf{r}}$, 
resulting in the transformed teacher predictions $\tilde{\mathbf{Y}}_T = \{\tilde{\mathbf{Y}}_T^L, \tilde{\mathbf{Y}}_T^U \}$.
With the same transformation, the dense matching between the two sets of predictions is naturally established, which is further filtered to increase the quality of pseudo-labels. 
The student network is optimized by enforcing a consistency loss on the remaining match set and a supervised loss between the student predictions and the ground-truth labels. In the following sections, we describe three core components of the framework. The colors of the titles are the same as the corresponding regions in Fig.~\ref{fig:framework}.

\textbf{\textcolor{orange}{Aligning Teacher-Student Predictions.}}
Suppose $\mathbf{A} = (\hat{x}, \hat{y}, \hat{z})$ represents the coordinate of the voxel. Following \cite{rukhovich2022fcaf3d}, the teacher prediction $\mathbf{y}^{\mathbf{A}} \in {\mathbf{Y}}_T$ is formulated by the bounding box parameters $\mathbf{\boldsymbol\delta}^\mathbf{A} = \{\delta_i^{\mathbf{A}}\}_{i=1}^{8}$, the centerness $c^\mathbf{A} \in [0, 1]$ and the semantic regression scores $\{p_i^\mathbf{A}\}_{i=1}^{N_{\text{cls}}}$, where $N_{\text{cls}}$ denotes the number of semantic categories.
The first six bounding box parameters $\delta_1, \delta_2, ..., \delta_6$ represent the distance to the opposite surfaces of the bounding box in the width, length and height dimension, and $\delta_7, \delta_8$ utilize the topological equivalency of pair $(\frac{w}{l}, \theta)$ to a Mobius strip for the disambiguation of heading angles of symmetric objects, namely:
\begin{equation}
\label{eq:param}
\resizebox{.9\linewidth}{!}{
\begin{math}
\begin{gathered}
    \delta_1^\mathbf{A} = (x + \frac{w}{2}) - \hat{x},~ 
    \delta_2^\mathbf{A} = \hat{x} - (x - \frac{w}{2}),~ 
    \delta_3^\mathbf{A} = (y + \frac{l}{2}) - \hat{y},\\
    \delta_4^\mathbf{A} = \hat{y} - (y - \frac{l}{2}),~ 
    \delta_5^\mathbf{A} = (z + \frac{h}{2}) - \hat{z},
    \delta_6^\mathbf{A} =  \hat{z} - (z - \frac{h}{2}),\\
    \delta_7^\mathbf{A} = \log\frac{w}{l}\sin(2\theta),~
    \delta_8^\mathbf{A} = \log\frac{w}{l}\cos(2\theta)
\end{gathered}
\end{math}
}
\end{equation}

Assuming the transformation $\mathbf{R}_{\theta, \Delta \mathbf{r}}$ maps $\mathbf{y}^{\mathbf{A}}$ to $\tilde{\mathbf{y}}^{\mathbf{A}'} := \tilde{\mathbf{y}}^{\mathbf{A}\mathbf{R}_{\theta, \Delta \mathbf{r}}}$.
Since the rotation around the upright-axis and the spatial translation have no effect on the semantics or the relative location towards anchor voxel of the predicted bounding box, we have $\tilde{c}^{\mathbf{A}'} = c^{\mathbf{A}}, \tilde{\mathbf{s}}^{\mathbf{A}}=\mathbf{s}^{\mathbf{A}'}$.
The relationship between $\tilde{\boldsymbol\delta}^{\mathbf{A}'}$ and ${\boldsymbol\delta}^{\mathbf{A}}$ can be derived from Eq.~\ref{eq:param}, which goes:
\begin{equation}
\label{eq:deltas}
\resizebox{.9\linewidth}{!}{
\begin{math}
\begin{gathered}
    \tilde\delta_1^\mathbf{A'} = \frac{\cos \theta + 1}{2} \delta_1^\mathbf{A}
                                 + \frac{-\cos \theta + 1}{2} \delta_2^\mathbf{A}
                                 + \frac{-\sin\theta}{2} \delta_3^\mathbf{A}
                                 + \frac{\sin\theta}{2} \delta_4^\mathbf{A}, \\
    \tilde\delta_2^\mathbf{A'} = \frac{-\cos \theta + 1}{2} \delta_1^\mathbf{A}
                                 + \frac{\cos \theta + 1}{2} \delta_2^\mathbf{A}
                                 + \frac{\sin\theta}{2} \delta_3^\mathbf{A}
                                 + \frac{-\sin\theta}{2} \delta_4^\mathbf{A}, \\
    \tilde\delta_3^\mathbf{A'} = \frac{\sin\theta}{2} \delta_1^\mathbf{A} 
                                 + \frac{-\sin\theta}{2} \delta_2^\mathbf{A}
                                 + \frac{\cos \theta + 1}{2} \delta_3^\mathbf{A}
                                 + \frac{-\cos \theta + 1}{2} \delta_4^\mathbf{A}, \\
    \tilde\delta_4^\mathbf{A'} = \frac{-\sin\theta}{2} \delta_1^\mathbf{A} 
                                 + \frac{\sin\theta}{2} \delta_2^\mathbf{A}
                                 + \frac{-\cos \theta + 1}{2} \delta_3^\mathbf{A}
                                 + \frac{\cos \theta + 1}{2} \delta_4^\mathbf{A}, \\
    \tilde\delta_5^\mathbf{A'} = \delta_5^\mathbf{A},~
    \tilde\delta_6^\mathbf{A'} =  \delta_6^\mathbf{A},~
    \tilde\delta_7^\mathbf{A'} = \delta_7^\mathbf{A} \cos (2\theta),~
    \tilde\delta_8^\mathbf{A'} = \delta_8^\mathbf{A} \cos (2\theta).
\end{gathered}
\end{math}
}
\end{equation}

The detailed derivation is in the supplementary material.

\definecolor{Matching Filtering}{RGB}{73,182,209}
\textbf{\textcolor{Matching Filtering}{Matching Filtering.}}
After establishing the matching between the two sets of predictions, a filtering strategy based on confidence is applied to the matching to reduce low-quality supervision. 
Specifically, with the predicted centerness score and semantic distribution in teacher outputs denoted by  $\tilde{\mathbf{c}}_T$ and $\tilde{\mathbf{s}}_T$, only matching that satisfies $\tilde{\mathbf{c}}_T > \tau_{\text{center}}$ and $\max\left(\rm{softmax}\left(\tilde{\mathbf{s}}_T\right)\right) > \tau_{\text{cls}}$ is retained. $\tau_{\text{center}}$ and $\tau_{\text{cls}}$ are hyperparameters.
Note that, even after the filtering, the matching in our method is still dense.

The primary distinction between the proposed \textit{dense matching} method and prior arts with \textit{proposal matching} is the processing order of the matching and filtering. 
In proposal matching methods, teacher proposals are first filtered using confidence scores for higher quality and then matched, resulting in even sparser teacher proposals. 
In the dense matching framework, on the contrary, the matching is established first and then filtered, preserving the spatial alignment of the predictions.

\begin{figure}[htbp]
    \centering
    \includegraphics[width=0.85\linewidth]{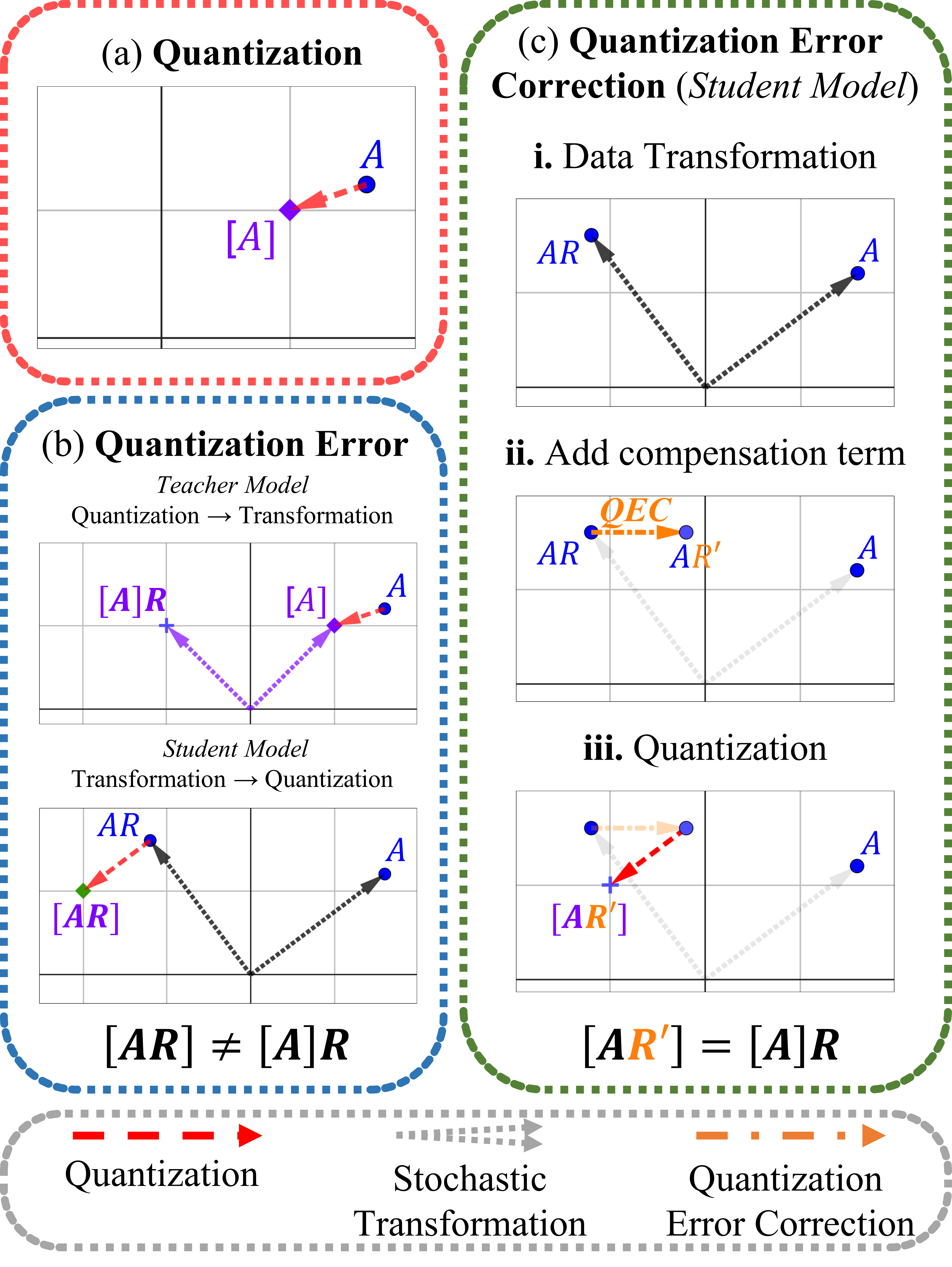}
    \caption{
    Demonstration of quantization error and correction. $R$ denotes stochastic augmentation and $[\cdot]$ denotes quantization. For the purpose of illustration, voxels and transformations are depicted in 2D space.
    (a) Concept of quantization.
    (b) Concept of quantization error. Without loss of generality, the random transformation is represented by a 90\degree counter-clockwise rotation. 
    (c) The process of quantization error correction (QEC). QEC is applied on the student branch between random transformation and quantization to eliminate the quantization error. 
    }
    \label{fig:qec}
    \vspace{-15pt}
\end{figure}

\definecolor{Losses}{RGB}{113,208,64}
\textbf{\textcolor{Losses}{Optimization.}}
The student model is optimized by gradient descent with the supervised loss $\mathcal{L}_{\text{supervised}}$ and the consistency loss $\mathcal{L}_{\text{consistency}}$.

The supervised loss $\mathcal{L}_{\text{supervised}}$ is enforced between the student predictions of the labeled input point clouds $\{\tilde{\mathbf{Y}}^L_S\}$ and the corresponding labels after the augmentation $\{\tilde{\mathbf{Y}}^L\}$.
Following \cite{rukhovich2022fcaf3d}, we adopt 3DIoU loss on the predicted bounding boxes, a binary cross entropy loss on the predicted centerness and a cross entropy loss on the predicted semantic distribution.

The consistency losses $\mathcal{L}_{\text{consistency}}$ are enforced on the filtered matching between student and teacher predictions.
For box parameters, we adopt Huber loss, which is less sensitive to outliers in pseudo labels:
\begin{equation}
    \resizebox{.9\linewidth}{!}{
    \begin{math}
    \begin{aligned}
    \mathcal{L}_{\text{box}}^{\mathbf{A}} = \begin{cases}
        \frac{1}{2}(\Delta \boldsymbol\delta^{\mathbf{A}}_i)^2, & \text{for} \ |\Delta \boldsymbol\delta^{\mathbf{A}}_i| < \tau_{\text{box}}, \\
        \tau_{\text{box}} \cdot (|\Delta \boldsymbol\delta^{\mathbf{A}}_i| - \frac{1}{2}\tau_{\text{box}}), & \text{otherwise}.
    \end{cases}
    \end{aligned}
    \end{math}
    }
\end{equation}
where $\Delta \boldsymbol\delta^{\mathbf{A}} = \boldsymbol\delta^{\mathbf{A}} - \tilde{\boldsymbol\delta}^{\mathbf{A}}$ and $\tau_{\text{box}}$ is a hyperparameter.
For centerness we adopt L2 loss $\mathcal{L}^{\mathbf{A}}_{\text{center}} = || c^{\mathbf{A}} - \tilde c^{\mathbf{A}} ||_2^2$.
For predicted semantic distribution, we adopt KL divergence $\mathcal{L}^{\mathbf{A}}_{\text{semantic}} = \sum_{c=1}^{N_{\text{cls}} } \mathbf{s}^{\mathbf{A}}_c \log (\frac{\mathbf{s}_c^{\mathbf{A}}}{\tilde{\mathbf{s}}_c^{\mathbf{A}}})$.
The final consistency loss is then formulated as:
\begin{equation}
\resizebox{.9\linewidth}{!}{
\begin{math}
\begin{aligned}
    \mathcal{L}_{\text{consistency}} = \lambda_{\text{box}} \mathcal{L}_\text{box} + \lambda_\text{center} \mathcal{L}_\text{center} + \lambda_\text{semantic} \mathcal{L}_\text{semantic}.
\end{aligned}
\end{math}
}
\end{equation}
where $\lambda_{\text{box}}$, $\lambda_\text{center}$ and $\lambda_\text{semantic}$ are loss weights.

As for the teacher model, the gradients are detached and the model parameters are updated using exponential moving average (EMA) of those of the student model:
\begin{equation}
    \theta_{t}^{n+1} = \alpha\theta_{t}^{n} + (1 - \alpha) \theta_{s}^{n}
\end{equation}
where $\theta_{t}^n$ and $\theta_{s}^n$ denote the parameters of the teacher and student networks at the $n$-th step, and $\alpha$ is the average factor.
The quality of the guidance provided by the teacher model is gradually improved with the knowledge from the student model.

\subsection{Quantization Error Correction}
\label{quantization}
In this section, we shed light on the problem of quantization error and propose a quantization error correction (QEC) module with closed-form solutions to address the problem.

Following implementation in MinkowskiEngine \cite{choy20194d}, we define the quantization (or voxelization) operator $[\cdot]$ on a vector as $[\mathbf{A}] = (\left\lfloor x_\mathbf{A}\right\rfloor, \left\lfloor y_\mathbf{A}\right\rfloor, \left\lfloor z_\mathbf{A}\right\rfloor )$, where the notation $\left\lfloor \cdot\right\rfloor$ denotes the floor operator.
The process of quantization is illustrated in Fig.~\ref{fig:qec}(a).
Since the stochastic transformation does not commute with voxelization (depicted in Fig.~\ref{fig:qec}(b)), the spatial location of an input point corresponds to two different ones after being processed by the student and teacher networks (\textcolor[rgb]{0.5, 0, 1.0}{$[\mathbf{AR}]$} and \textcolor[rgb]{0.5, 0, 1.0}{$[\mathbf{A}]\mathbf{R}$} in Fig.~\ref{fig:qec}(b)), the difference of which is defined as the \textbf{\textit{quantization error}}.

The quantization error is detrimental to dense pseudo-label self-training scheme, as it violates the exact dense matching we pursue and causes inaccurate training signals and performance decrease.
We propose an online solution that finds a compensation term $\vec{\mathbf{r}^{\prime}}(\mathbf{A}, \mathbf{R}_{\theta, \Delta \mathbf{r}})$ for the given location $\mathbf{A}$ and transformation $\mathbf{R}_{\theta, \Delta \mathbf{r}} \in \mathcal{T}$, namely find $\vec{\mathbf{r}'}$ that satisfies:
\begin{equation}
    \label{eq:samevoxel}
    [ \mathbf{A}\mathbf{R}_{\theta, \Delta \mathbf{r}}  + \vec{\mathbf{r}'} ] \xlongequal{\rm{same\ voxel}}{} [\mathbf{A}] \mathbf{R}_{\theta, \Delta \mathbf{r}}
\end{equation}
We rewrite Eq.~\ref{eq:samevoxel} by applying voxelization to both sides to replace the \textit{same voxel} equality with the arithmetic equality. Since quantization operation holds the property of idempotence, we have:
\begin{equation}
    \label{eq:eqv}
    [ \mathbf{A}\mathbf{R}_{\theta, \Delta \mathbf{r}}  + \vec{\mathbf{r}'} ] = [ [\mathbf{A}] \mathbf{R}_{\theta, \Delta \mathbf{r}}]
\end{equation}
By refactoring $\mathbf{A}\mathbf{R}_{\theta, \Delta \mathbf{r}}$ into $\mathbf{A}\mathbf{R}_{\theta} + \Delta \mathbf{r}$ and using \textit{Lemma.1} and \textit{Lemma.2} from the supplementary material, we have:
\begin{equation}
    \label{eq:equal0}
    [\{\mathbf{A}\} \mathbf{R}_{\theta} + \{\Delta \mathbf{r}\} + \vec{\mathbf{r}'}] = \mathbf{0}
\end{equation}
when $\theta \in \{ \frac{k\pi}{2} \}_{k=0}^{3}$.
The operator $\{ \cdot \} : \mathbf{x} \mapsto \mathbf{x} - [\mathbf{x}]$ is defined as the remainder after quantization.
We solve Eq.~\ref{eq:equal0} by interpreting the equation as a requirement for the terms on the left-hand side to lie within the voxel represented by the original point.
Therefore, assuming $\boldsymbol\gamma \in [0, S_\text{v}]^3$ ($S_\text{v}$ denotes the voxel size), we derive the compensation term for $\vec{\mathbf{r}^{\prime}}(\mathbf{A}, \mathbf{R}_{\theta, \Delta \mathbf{r}})$ as:
\begin{equation}
    \label{eq:closed-form}
    \vec{\mathbf{r}'}(\boldsymbol\gamma) = \boldsymbol\gamma - \{\mathbf{A}\} \mathbf{R}_{\theta} - \{\Delta \mathbf{r}\}
\end{equation}
To alleviate the negative impacts caused by the perturbations to the point cloud structures, we select $\boldsymbol\gamma_0$ such that:
\begin{equation}
    \label{eq:final-eq}
    \boldsymbol\gamma_0 = \rm{argmin}_{\boldsymbol\gamma\in [0, S_\text{v}]^3} \ || \boldsymbol\gamma - \{\mathbf{A}\} \mathbf{R}_{\theta} - \{\Delta \mathbf{r}\}  ||_2
\end{equation}
Finding $\boldsymbol\gamma_0$ is a typical mathematical optimization problem, 
 and we provide a closed-form solution to Eq.~\ref{eq:final-eq} in the supplementary material.


\section{Experiments}

\subsection{Datasets}
Following prior arts \cite{wang20213dioumatch}\cite{zhao2020sess} aiming at semi-supervised 3D object detection, we evaluate our framework on ScanNet v2 \cite{dai2017scannet} and SUN RGB-D \cite{song2015sun}.

\textbf{ScanNet v2} \cite{dai2017scannet} is a widely used 3D indoor scene dataset which contains 1512 scans of indoor scenes reconstructed from 2.5 million high-quality RGB-D images.
The annotations include per-point instance labels which enable the derivation of axis-aligned object bounding boxes for training and evaluation of 3D object detection methods. 
The challenge with this dataset in the semi-supervised setting is the limited amount of labeled data. 
For instance, the 5\% labeled setting corresponds to only a few dozen labeled scenes, making it difficult to learn a good detector from labeled data alone. 

\textbf{SUN RGB-D} \cite{song2015sun} is a widely used benchmark dataset with 10335 indoor scene scans for evaluating scene understanding algorithms, particularly in the context of 3D object detection. Apart from the RGB-D data, the dataset also provides ground-truth 3D bounding box annotations, which enables the evaluation of the task of 3D object detection.
The main challenge of this dataset is that the scenes are not axis-aligned. This rotational variability makes it difficult to predict object bounding boxes accurately in the semi-supervised setting, as the model is challenged to recognize objects with any possible orientation after training on limited labeled data.

\subsection{Implementation Details}

\textbf{Hyperparameters.} We use the same set of hyperparameters for both datasets.
As suggested in \cite{rukhovich2022fcaf3d}, the voxel size is set to $S_\text{v} = 0.01\text{m}$.
The confidence thresholds are set to $\tau_\text{center} = 0.40$ and $\tau_\text{cls} = 0.20$.
The threshold for Huber loss is set to $\tau_\text{box}=0.30$.
The weights for consistency losses are set to $\lambda_{\text{box}}=1.00$, $\lambda_{\text{center}}=0.25$, and $\lambda_{\text{semantic}}=0.50$.
The same warmup strategy as in \cite{zhao2020sess} is adopted for the consistency losses.
The average factor $\alpha$ of the exponential moving average is set to $0.999$.
As for the stochastic transformation strategies, rotation $\theta$ around the upright-axis is randomly chosen from $\{0, \frac{\pi}{2}, \pi, \frac{3\pi}{2} \}$ and random translation $\Delta \mathbf{r}$ is sampled uniformly from $[-0.5\text{m}, 0.5\text{m}]^3$.

\textbf{Details during training and evaluation.} 
We use MMDetection3D \cite{mmdet3d2020} toolbox to implement our proposed framework.
For semi-supervised detection on both ScanNet and SUN RGB-D, our method runs for 12000 training steps which empirically leads to good convergence.
During training, we adopt the AdamW optimizer \cite{loshchilov2017decoupled} with an initial learning rate of $10^{-3}$ and a weight decay factor of $10^{-4}$, and a scheduler decaying the learning rate by 90\% at 67\% and 90\% of the training process.
In the semi-supervised setting, a training batch contains 8 labeled samples and 8 unlabeled samples.
During evaluation, to ensure fair comparison with former semi-supervised 3D object detection methods \cite{zhao2020sess, wang20213dioumatch}, we perform only one forward pass without test-time augmentation used by \cite{rukhovich2022fcaf3d}. Meanwhile, we keep other evaluation settings including IoU thresholds for NMS the same as \cite{rukhovich2022fcaf3d}. We report the $\rm{mAP@0.25}$ and $\rm{mAP@0.50}$ metrics.

\subsection{Comparison of Matching Schemes}
\label{comparison}
In this section, we first provide the comparison between our proposed \textbf{\textit{dense matching}} method and the \textbf{\textit{proposal matching}} methods.
We validate the superiority of our proposed methods by demonstrating the quantity and quality of the pseudo labels generated by dense matching strategy.


\begin{figure}[tbp]
    \centering
    \includegraphics[width=0.85\linewidth]{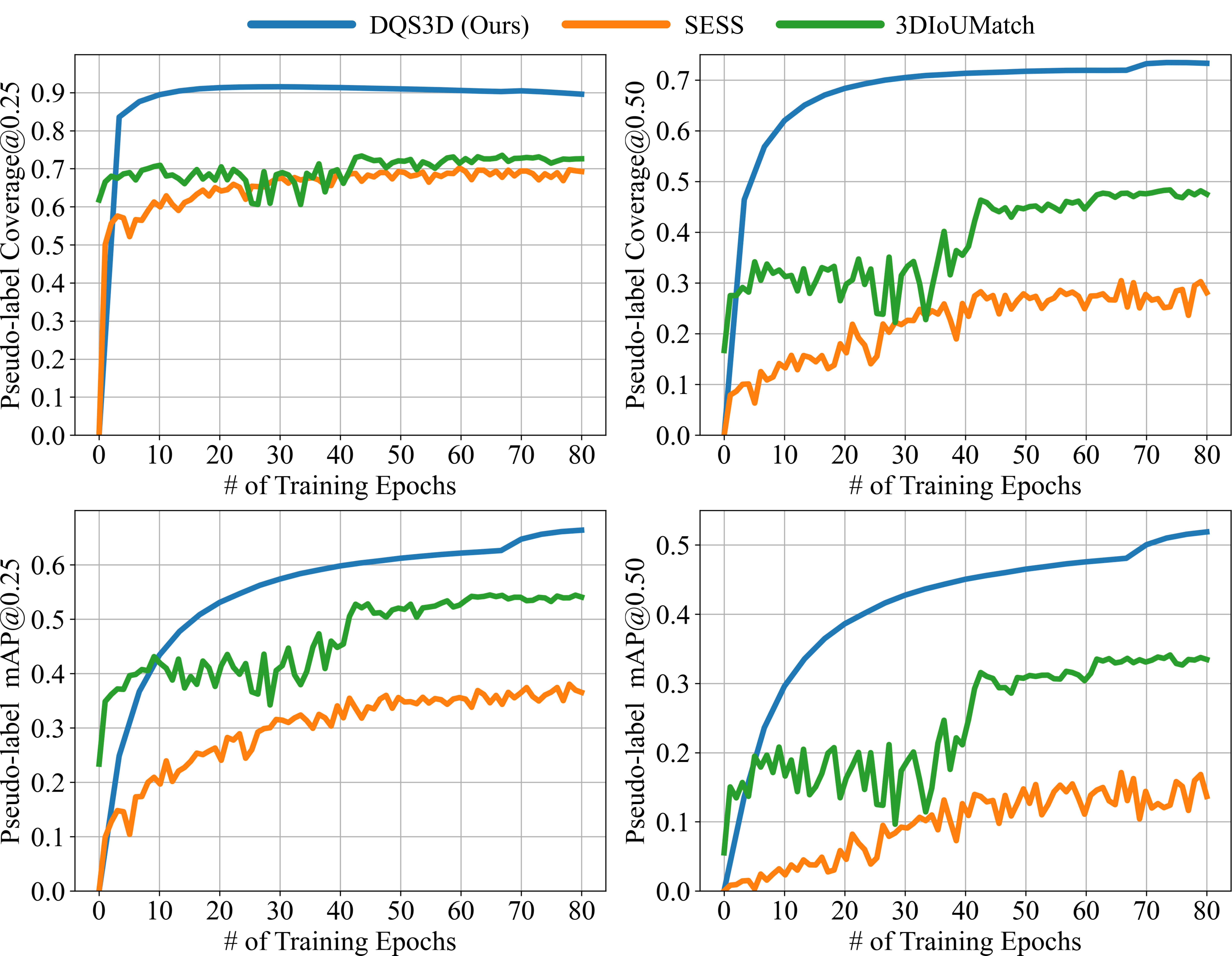}
    \caption{Transductive analysis on different matching schemes in semi-supervised 3D object detection. Experiments are conducted on ScanNet with 10\% training data equipped with labels. $\rm{Coverage}@\{0.25, 0.50\}$ and $\rm{mAP}@\{0.25, 0.50\}$ on the unlabeled set are reported. The proposed dense matching scheme achieves significantly higher performance than prior arts.
    }
    
    \label{fig:my_label}
    \vspace{-15pt}
\end{figure}

\begin{table*}
\centering
\resizebox{0.75\linewidth}{!}{
\begin{tblr}{
  cells = {c,},
  cell{1}{1} = {r=2}{},
  cell{1}{2} = {r=2}{},
  cell{1}{3} = {c=2}{},
  cell{1}{5} = {c=2}{},
  cell{1}{7} = {c=2}{},
  cell{1}{9} = {c=2}{},
  cell{3-4}{3-8} = {fg=gray8},
  cell{9-10}{3-8} = {fg=gray8},
  cell{5-6}{9-Z} = {fg=gray8},
  cell{11-12}{9-Z} = {fg=gray8},
  cell{3}{1} = {r=6}{white},
  cell{9}{1} = {r=6}{white},
  row{1-2} = {},
  hline{1, 2, 3,9,15}={},
  hline{7, 13}={2-Z}{},
  vline{2,3, 9}={}
}
  & Model        & 5\%           &               & 10\%          &               & 20\%          &               & 100\%         &               \\
         &              & {\rm{mAP}
\\\rm{@0.25}} & {\rm{mAP}
\\\rm{@0.50}} & {\rm{mAP}
\\\rm{@0.25}} & {\rm{mAP}
\\\rm{@0.50}} & {\rm{mAP}
\\\rm{@0.25}} & {\rm{mAP}
\\\rm{@0.50}} & {\rm{mAP}
\\\rm{@0.25}} & {\rm{mAP}
\\\rm{@0.50}} \\
\rotatebox{90}{ScanNet \cite{dai2017scannet}} & VoteNet \cite{qi2019deep}      & 27.9          & 10.8          & 36.9          & 18.2          & 46.9          & 27.5          & 57.8          & 36.0          \\
         & FCAF3D  \cite{rukhovich2022fcaf3d}      & 43.8 & 29.3 & 51.1 & 35.7 &58.2 & 42.1 &69.5 & 55.1 \\
         & SESS \cite{zhao2020sess}         & 32.0          & 14.4          & 39.5          & 19.8          & 49.6          & 29.0          & 61.3          & 39.0          \\
         & 3DIoUMatch \cite{wang20213dioumatch}  & 40.0          & 22.5          & 47.2          & 28.3          & 52.8          & 35.2          & 62.9          & 42.1          \\
         & \textbf{DQS3D (Ours)} &         \textbf{49.2}      &        \textbf{35.0}       &    \textbf{57.1}           &       \textbf{41.8}        &        \textbf{64.3}       &       \textbf{48.5}        &          \textbf{71.9}     &        \textbf{56.3}       \\
         & Improv. & \textcolor{blue}{\textbf{+9.2~$\uparrow$}} & \textcolor{blue}{\textbf{+12.5~$\uparrow$}} & \textcolor{blue}{\textbf{+9.9~$\uparrow$}} & \textcolor{blue}{\textbf{+13.5~$\uparrow$}} & \textcolor{blue}{\textbf{+11.5~$\uparrow$}} & \textcolor{blue}{\textbf{+13.3~$\uparrow$}} & \textcolor{magenta}{\textbf{+2.4~$\uparrow$}} & \textcolor{magenta}{\textbf{+1.2~$\uparrow$}}  \\
\rotatebox{90}{SUN-RGBD \cite{song2015sun}} & VoteNet \cite{qi2019deep}      & 29.9          & 10.5          & 38.9          & 17.2          & 45.7          & 22.5          & 58.0          & 33.4          \\
         & FCAF3D  \cite{rukhovich2022fcaf3d}     &     49.5          &          31.7     &             50.7  &      33.4         &       54.3        & 36.5              &   63.6            &          47.5     \\
         & SESS   \cite{zhao2020sess}      & 34.2          & 13.1          & 42.1          & 20.9          & 47.1          & 24.5          & 60.5          & 38.1          \\
         & 3DIoUMatch \cite{wang20213dioumatch}  & 39.0          & 21.1          & 45.5          & 28.8          & 49.7          & 30.9          & 61.5          & 41.3          \\
         & \textbf{DQS3D (Ours)} &     \textbf{53.2}          &         \textbf{35.6}      &   \textbf{55.7}            &          \textbf{38.2}     &      \textbf{58.0}         &            \textbf{42.3}   &       \textbf{64.1}        &        \textbf{48.2}    \\
         & Improv. & \textcolor{blue}{\textbf{+14.2~$\uparrow$}} & \textcolor{blue}{\textbf{+14.5~$\uparrow$}} & \textcolor{blue}{\textbf{+10.2~$\uparrow$}} & \textcolor{blue}{\textbf{+9.4~$\uparrow$}} & \textcolor{blue}{\textbf{+8.3~$\uparrow$}} & \textcolor{blue}{\textbf{+11.4~$\uparrow$}} & \textcolor{magenta}{\textbf{+0.5~$\uparrow$}} & \textcolor{magenta}{\textbf{+0.7~$\uparrow$}}  \\
\end{tblr}
}

\caption{Experiment results on the task of 3D object detection in various semi-supervised settings (5\%, 10\%, 20\% labels available) and the fully-supervised setting on ScanNet \cite{dai2017scannet} and SUN-RGBD \cite{song2015sun} datasets.
The proposed DQS3D is compared with \textbf{semi-supervised} 3D object detection frameworks SESS \cite{zhao2020sess} and 3DIoUMatch \cite{wang20213dioumatch}, with the margins over 3DIoUMatch \cite{wang20213dioumatch} marked in \textcolor{blue}{blue}.
Proficient Teachers \cite{yin2022semi} is currently not comparable as their experiments were only conducted in outdoor scenes and their codes are not currently available, which hinders us to reproduce their experiments on indoor benchmarks.
DQS3D is also compared with \textbf{fully-supervised} 3D object detectors VoteNet \cite{qi2019deep} and FCAF3D \cite{rukhovich2022fcaf3d}, with the margins over FCAF3D \cite{rukhovich2022fcaf3d} marked in \textcolor{magenta}{magenta}.
}
\label{table:main}
\vspace{-10pt}
\end{table*}

\textbf{Are more pseudo-labels harvested?}
We trained our proposed method as well as former arts on ScanNet \cite{dai2017scannet} dataset with 10\% data equipped with labels and collect the pseudo labels harvested by these methods.
The average amounts of pseudo-labels harvested from one scene are illustrated in Fig.~\ref{fig:teaser}.
As shown in the figure, our methods with dense matching strategy harvest a significantly larger amount of pseudo labels compared with SESS \cite{zhao2020sess}, 3DIoUMatch \cite{wang20213dioumatch} and Proficient Teachers \cite{yin2022semi}, which is a contributing factor to better performance for semi-supervised 3D object detection.
As shown by later experiments in Sec.~\ref{exp:sotas}, this translates to the improvement of the detection performance.

\textbf{Are the pseudo-labels of good quality?}
In this section, we investigate the quality of pseudo-labels generated by different matching schemes.
We borrow the concept of transductive analysis \cite{zhao2020sess} where we regard the model performance on the unlabeled set as the indicating measure of the pseudo-label quality.
The results are obtained by training our methods and former arts on ScanNet \cite{dai2017scannet} dataset with 10\% of data equipped with labels.
In Fig.~\ref{fig:my_label}, we depict the $\rm{Coverage@\{0.25, 0.50\}}$ and $\rm{mAP@\{0.25, 0.50\}}$ on the unlabeled set during the training stage, where $\rm{Coverage}$ indicates the recall rate of objects in the scene.
It can be observed that dense matching provides significantly more informative and accurate pseudo-labels, compared with the proposal-based counterparts.
We attribute the improved transductive results to the way in which the dense prediction scheme resolves the "\textcolor[rgb]{0.9257, 0.5, 0.1914}{multiple supervision}" and "\textcolor[rgb]{0.4375 0.1875 0.625}{no supervision}" issues demonstrated in Fig.~\ref{fig:dense-matching}(a) and provides spatially dense training signals for the student network. For more evidence, see Fig.~\ref{fig:viz_result}.

\subsection{Comparisons with prior SOTAs}
\label{exp:sotas}
In this section, we conduct extensive experiments and report the performance of DQS3D and the prior arts in both semi-supervised and fully-supervised settings on the ScanNet and SUN RGB-D datasets. 
In the semi-supervised setting, the proportion of the labeled set varies among 5\%, 10\%, and 20\%.
The consistency losses are imposed on both the labeled and unlabeled sets.
In the fully-supervised setting, the entire dataset is regarded as both the labeled and unlabeled sets to examine if the proposed framework can further learn from the additional supervision of pseudo-labels. The experiment results are presented in Tab.~\ref{table:main}.

Our method outperforms prior proposal matching methods by large margins and sets new state-of-the-art results on the semi-supervised 3D object detection benchmark for both ScanNet and SUN RGB-D datasets.
It is noteworthy that the improvements of $\rm{mAP@0.50}$ are generally larger than those of $\rm{mAP@0.25}$.
We attribute this to the denseness of the pseudo labels, which provides more spatially fine-grained supervision for the student model.
In this way, the predicted object bounding boxes overlap with the target objects to a larger extent, which helps achieve more distinct margins with higher IoU thresholds.

Surprisingly, in the fully-supervised setting, our method also pushes the boundaries of 3D object detection, as shown in Tab.~\ref{table:main}.
We attribute these improvements to the way in which the framework of self-training serves as regularization and helps improve the stability of the training procedure, as the pseudo-labels generated by the teacher networks are not affected by the bias in mini-batches.


\subsection{Ablation Studies}

\textbf{Quantization Error Correction.}
To demonstrate effectiveness of our proposed quantization error correction (QEC) module (detailed in Sec.~\ref{quantization}), we conduct experiments on ScanNet (20\% of training set equipped with labels), in which we train the proposed framework with various voxel sizes and ablate on the compensation term. 
In addition to the $\rm{mAP@\{0.25, 0.50\}}$ on the \textit{validation} set, we also report the weighted IoU of the predicted and ground-truth bounding boxes on the \textit{labeled training} set. 
The object's weight is determined by the predicted centerness. 

The results are presented in Tab.~\ref{tab:qec-ablation}.
Notably, with all voxel sizes, experiments trained with the compensation term achieve higher performances (up to \textcolor{blue}{+2.49\%} IoU) than those trained without the term.
The non-trivial improvements demonstrate the effectiveness of the QEC module in addressing the inherent issue of quantization error and consequently improving the detection accuracy.

\begin{figure}[htbp]
    \centering
    \begin{subfigure} {0.45\linewidth}
        \includegraphics[width=0.90\linewidth]{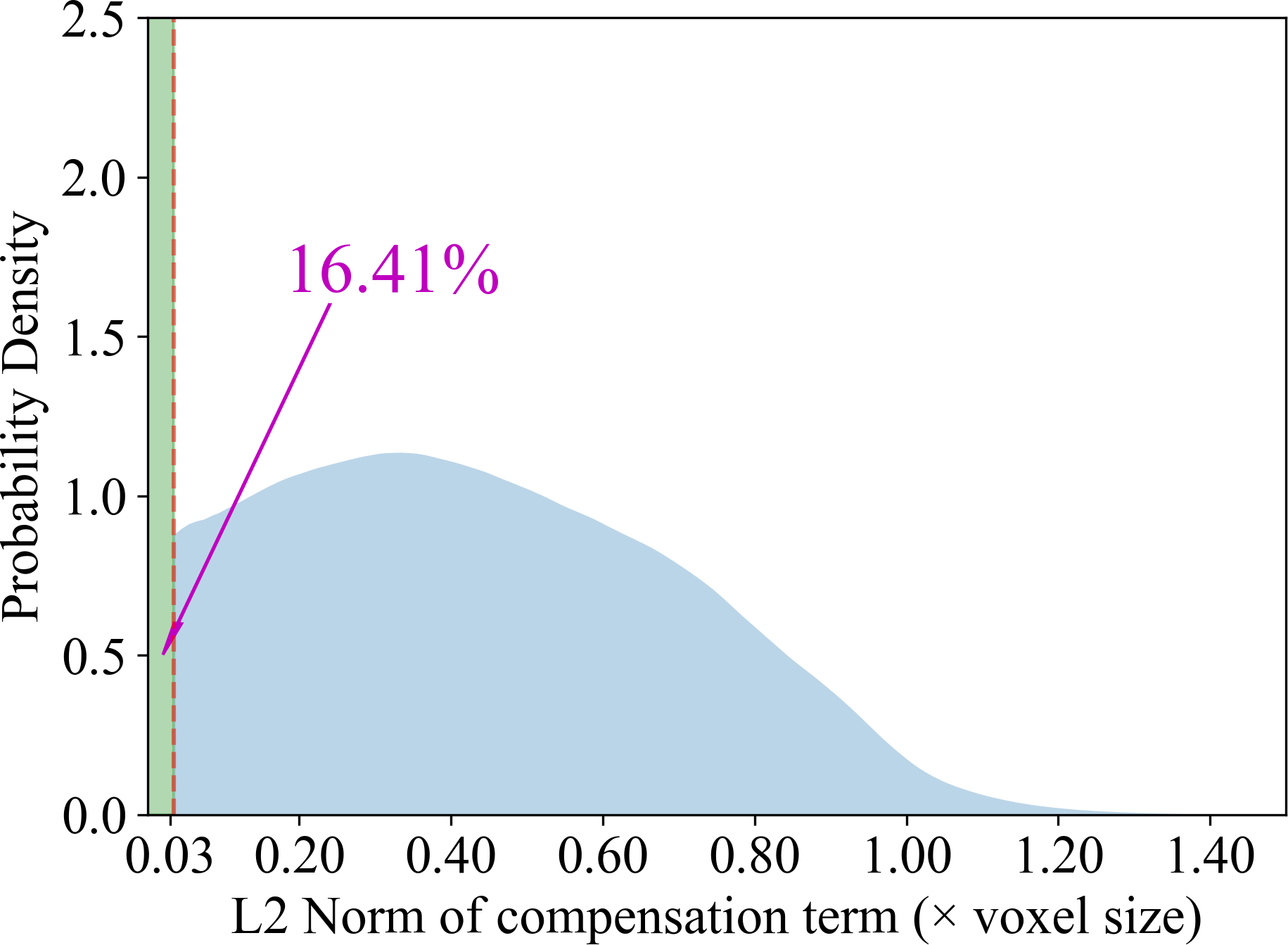}
    \end{subfigure}
    \hfill
    \begin{subfigure} {0.45\linewidth}
        \includegraphics[width=0.90\linewidth]{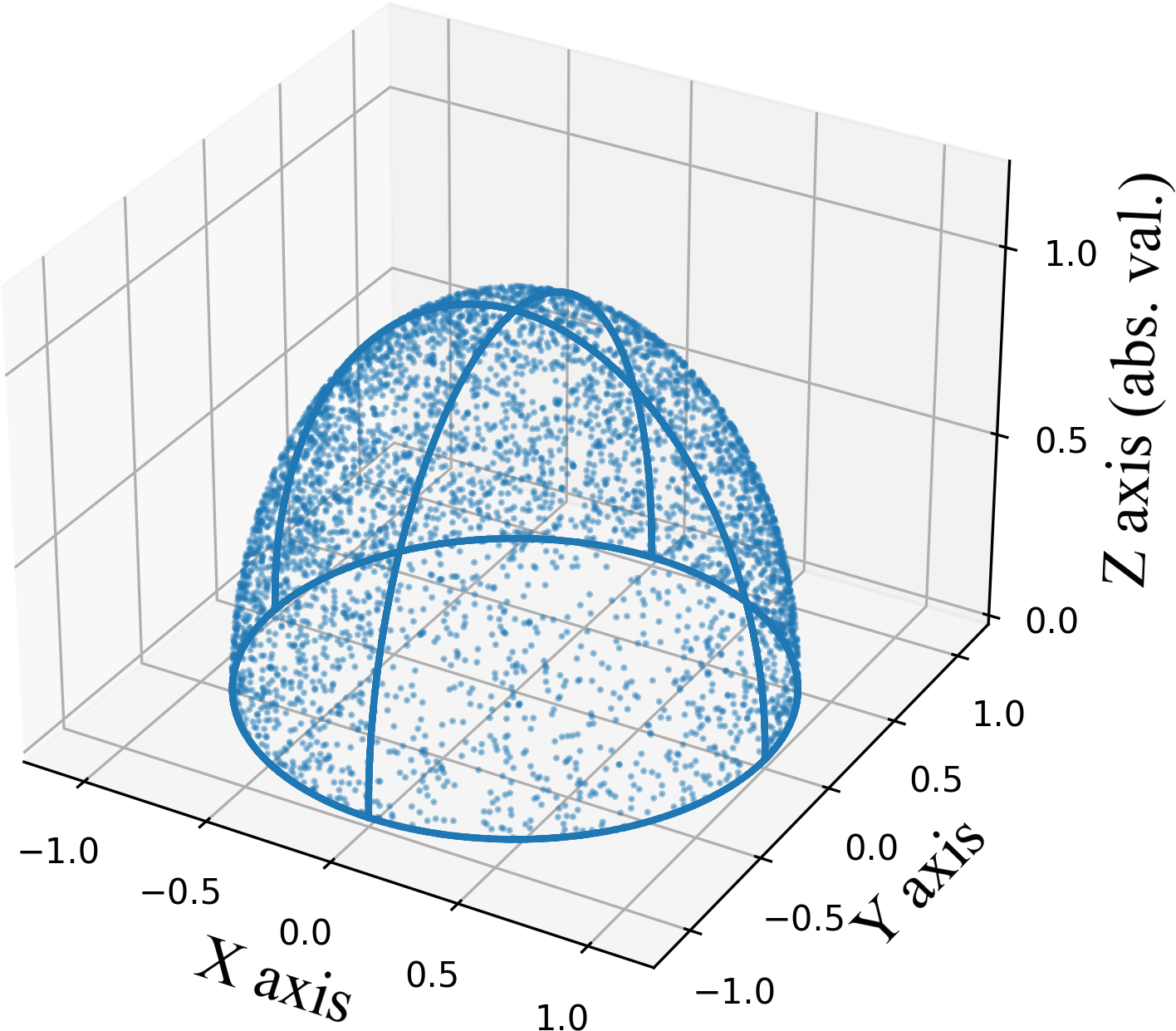}
    \end{subfigure}
    \caption{
    Visualized statistics of the Quantization Error Correction terms. QEC terms are collected from 80M points from ScanNet scenes under training-stage transformations.
    (a) L2 norm distribution of QEC terms.
    (b) Directions of QEC terms, note that \textbf{the solid blue lines are actually formed by a large amount of crowded data points.}
}
    \label{fig:qec_ablation}
    \vspace{-10pt}
\end{figure}

Furthermore, we conduct a statistical analysis for intuitively understanding the QEC module. 
In experiments training on ScanNet dataset, we collect the compensation terms of 80 million points and plot the distribution of the L2 norms and the directions of the terms in Fig.~\ref{fig:qec_ablation}.
As depicted in Fig.~\ref{fig:qec_ablation}(a), the L2 norms of more than 80\% of the compensation terms lie in the range of $[0.03 S_\text{v}, S_\text{v}]$ ($S_\text{v}$ denotes the voxel size), demonstrating that the quantization error is a non-trivial phenomenon.
As depicted in Fig.~\ref{fig:qec_ablation}(b), the majority of the compensation terms are aligned with axes. This is because the QEC terms have the smallest possible magnitude by design to preserve the point cloud structure.

\begin{figure}[tbp]
    \centering
    \includegraphics[width=0.80\linewidth]{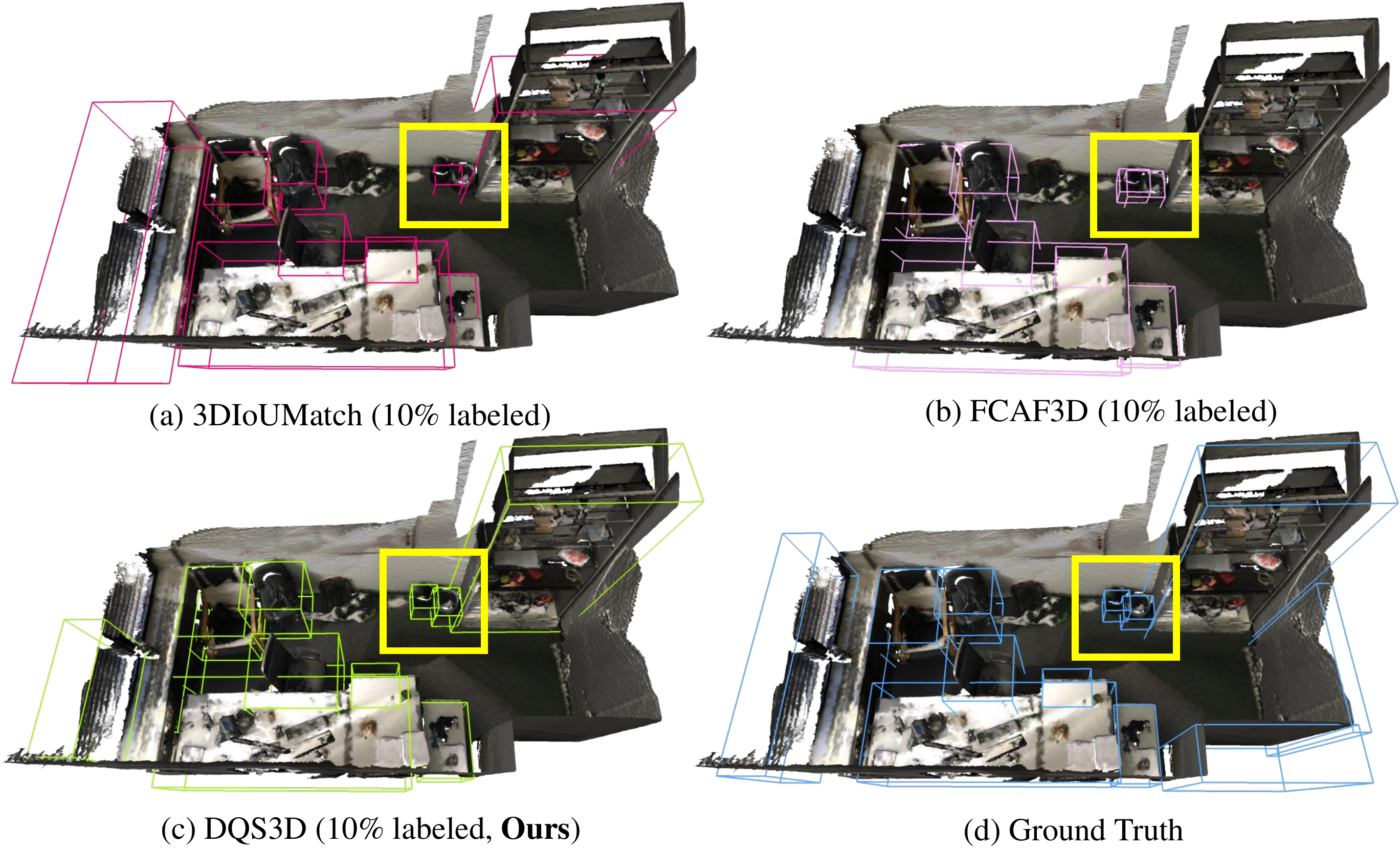}
    \caption{
Qualitative results for the 10\% labeled setting. Notice the two trash cans against the back wall. As for \textit{(a) 3DIoUMatch} \cite{wang20213dioumatch}, which employs proposal matching, a \textcolor[rgb]{0.9257, 0.5, 0.1914}{\textit{multiple supervision}} problem arises, because when adjacent teacher proposals align with the same student proposal, the student receives noisy supervision. However, \textit{(c) our dense matching} successfully resolves this issue.
    }
    \label{fig:viz_result}
    \vspace{-15pt}
\end{figure}

\begin{table}[htbp]
\begin{center}
\resizebox{0.9\columnwidth}{!}{
\begin{tblr}{
  cells = {c},
  cell{2}{1} = {r=2}{},
  cell{4}{1} = {r=2}{},
  cell{6}{1} = {r=2}{},
  hline{1,2,4,6,8,10}={},
  vline{2,3}={},
}
 Voxel Size (m) & QEC & IoU (\%) &  $\rm mAP@0.25$ & $\rm mAP@0.50$ \\
0.01 &            & 75.19 \phantom{\dt{+1.04}}                   & 63.7 \phantom{\dt{+1.4}}                  & 47.1 \phantom{\dt{+1.4}}                  \\ 
0.01 & \checkmark & \textbf{76.82} \textcolor{blue}{\dt{+1.63}}  & \textbf{64.3} \textcolor{blue}{\dt{+0.6}} & \textbf{48.5} \textcolor{blue}{\dt{+1.4}} \\ 
0.02 &            & 71.06 \phantom{\dt{+1.04}}                   & 59.0 \phantom{\dt{+1.4}}                  & 42.2 \phantom{\dt{+1.4}}                  \\ 
0.02 & \checkmark & \textbf{72.63} \textcolor{blue}{\dt{+1.57}}  & \textbf{60.2} \textcolor{blue}{\dt{+1.2}} & \textbf{42.8} \textcolor{blue}{\dt{+0.6}} \\ 
0.03 &            & 65.63 \phantom{\dt{+1.04}}                   & 51.9 \phantom{\dt{+1.0}}                  & 35.2 \phantom{\dt{+1.0}}                  \\ 
0.03 & \checkmark & \textbf{68.12} \textcolor{blue}{\dt{+2.49}}  & \textbf{52.7} \textcolor{blue}{\dt{+0.8}} & \textbf{35.9} \textcolor{blue}{\dt{+0.7}}\\ 
\end{tblr}
}
\end{center}
\vspace{-15pt}
\caption{Ablations on the QEC term. Experiments are conducted on ScanNet with 20\% training data with labels.}
\label{tab:qec-ablation}
\end{table}

\begin{table}[htbp]
\begin{center}
\resizebox{1.0\columnwidth}{!}{
\begin{tblr}{
  cells = {c},
  hline{1,2,3,6,9,11}={},
  vline{2,5}={},
  cell{2}{1-Z} = {bg=bgred},
  cell{9}{1-Z} = {bg=bgblue},
}
 T & Box & Centerness & Class & $\rm mAP@0.25$ & $\rm mAP@0.50$ \\
  \checkmark & \checkmark & \checkmark & \checkmark & \textbf{64.3} \phantom{\dt{-0.0}} \textcolor{blue}{\dt{+4.3}}& \textbf{48.5} \phantom{\dt{-0.0}} \textcolor{blue}{\dt{+3.9}}\\
   \checkmark &  & \checkmark & \checkmark & 62.3 \textcolor{red}{\dt{-2.0}} \textcolor{blue}{\dt{+2.3}} & 45.3 \textcolor{red}{\dt{-3.2}} \textcolor{blue}{\dt{+0.7}} \\
  \checkmark & \checkmark &  & \checkmark & 63.5 \textcolor{red}{\dt{-0.8}} \textcolor{blue}{\dt{+3.5}} & 46.4 \textcolor{red}{\dt{-2.1}} \textcolor{blue}{\dt{+1.8}}\\
  \checkmark & \checkmark & \checkmark &  &  63.4 \textcolor{red}{\dt{-0.9}} \textcolor{blue}{\dt{+3.4}}& 47.3 \textcolor{red}{\dt{-1.2}} \textcolor{blue}{\dt{+2.7}}\\
    \checkmark & \checkmark &  &  & 63.0 \textcolor{red}{\dt{-1.3}} \textcolor{blue}{\dt{+3.0}}& 46.3 \textcolor{red}{\dt{-2.2}} \textcolor{blue}{\dt{+1.7}}\\
   \checkmark &  & \checkmark &  & 61.6 \textcolor{red}{\dt{-2.7}} \textcolor{blue}{\dt{+1.6}} & 44.9 \textcolor{red}{\dt{-3.6}} \textcolor{blue}{\dt{+0.3}}\\
   \checkmark &  &   & \checkmark & 62.1 \textcolor{red}{\dt{-2.2}} \textcolor{blue}{\dt{+2.1}}&  45.3 \textcolor{red}{\dt{-3.2}} \textcolor{blue}{\dt{+0.7}}\\
    \checkmark  &   &  &  & 60.0 \textcolor{red}{\dt{-4.3}}\phantom{\dt{+4.3}} & 44.6 \textcolor{red}{\dt{-3.9}} \phantom{\dt{+4.3}}\\
      &   &  &  & 58.2 \textcolor{red}{\dt{-6.1}} \textcolor{blue}{\dt{-1.8}} & 42.1 \textcolor{red}{\dt{-6.4}} \textcolor{blue}{\dt{-2.5}}\\
\end{tblr}
}
\end{center}
\vspace{-15pt}
\caption{Ablations on the consistency losses. Experiments are conducted on ScanNet with 20\% training data equipped with labels. T denotes random transformations.
\textcolor{red}{Red margins} are comparisons with DQS3D with all consistency losses. \textcolor{blue}{Blue margins} are comparisons with DQS3D with no consistency losses.
}
\label{tab:consistency}
\vspace{-15pt}
\end{table}

\textbf{Consistency Losses.} 
Ablation experiments on the consistency losses are conducted on the ScanNet dataset with 20\% training data equipped with labels.
The results are reported in Tab.~\ref{tab:consistency}.
According to the results, the absence of the box consistency loss has the largest influence with the largest performance drops of \textcolor{red}{-2.0\%} ($\rm mAP@0.25$) and \textcolor{red}{-3.2\%} ($\rm mAP@0.50$), while the absences of other two consistency losses also bring about drops in performance. These results indicate that each component of the proposed consistency losses is necessary.


\section{Conclusion}
This paper presents a densely-matched quantization-aware framework, DQS3D, for semi-supervised 3D object detection. 
By leveraging dense matching instead of proposal matching, and by addressing the issue of quantization error, DQS3D achieves significant improvements over former arts on two widely-used benchmarks, ScanNet v2 and SUN RGB-D, in the semi-supervised setting. 

Furthermore, the paper provides evidence that the use of dense predictions leads to more meaningful pseudo-labels and promotes self-training. 
We hope the insights and techniques introduced in this work would inspire future research in the field of semi-supervised learning.\footnote{This work is sponsored by DiDi GAIA research program.}
\newpage
{\small
\bibliographystyle{ieee_fullname}
\bibliography{egbib}
}

\clearpage
\newpage
\mbox{~}

\appendix

\section{Proof of Equations}

\paragraph{Lemma 1} If all elements in $\mathbf{A}$ are integers, then the following equation holds:
\begin{equation}
    [\mathbf{A} + \mathbf{B}] = \mathbf{A} + [\mathbf{B}]
\end{equation}
\label{lemma:1}

\textbf{Proof}: By definition. $\square$

\paragraph{Lemma 2} If all elements in $\mathbf{A}$ are integers and $\theta \in \{ \frac{k\pi}{2} \}_{k=0}^{3}$, then all elements in $\mathbf{A}\mathbf{R}_{\theta}$ are integers.
\label{lemma:2}

\textbf{Proof}: By considering the rotation matrix $\mathbf{R}_\frac{k\pi}{2}$ when $k =  0,1,2,3$. 
Note that $x' = x \cos \theta + y \sin \theta$, $y' = -x\sin\theta + y \cos\theta$ and $z'=z$.
When $k =  0,1,2,3$, $\sin\theta$ and $\cos\theta$ produces integer values.
According to the property of integer fields, $x', y'$ and $z'$ are also integers, which means all elements in $\mathbf{A}\mathbf{R}_{\theta}$ are integers. $\square$

\paragraph{Proof of Equation 2} Here we prove:
\begin{equation}
\label{eq:deltas}
\resizebox{.85\linewidth}{!}{
\begin{math}
\begin{gathered}
    \tilde\delta_1^\mathbf{A'} = \frac{\cos \theta + 1}{2} \delta_1^\mathbf{A}
                                 + \frac{-\cos \theta + 1}{2} \delta_2^\mathbf{A}
                                 + \frac{-\sin\theta}{2} \delta_3^\mathbf{A}
                                 + \frac{\sin\theta}{2} \delta_4^\mathbf{A}, \\
    \tilde\delta_2^\mathbf{A'} = \frac{-\cos \theta + 1}{2} \delta_1^\mathbf{A}
                                 + \frac{\cos \theta + 1}{2} \delta_2^\mathbf{A}
                                 + \frac{\sin\theta}{2} \delta_3^\mathbf{A}
                                 + \frac{-\sin\theta}{2} \delta_4^\mathbf{A}, \\
    \tilde\delta_3^\mathbf{A'} = \frac{\sin\theta}{2} \delta_1^\mathbf{A} 
                                 + \frac{-\sin\theta}{2} \delta_2^\mathbf{A}
                                 + \frac{\cos \theta + 1}{2} \delta_3^\mathbf{A}
                                 + \frac{-\cos \theta + 1}{2} \delta_4^\mathbf{A}, \\
    \tilde\delta_4^\mathbf{A'} = \frac{-\sin\theta}{2} \delta_1^\mathbf{A} 
                                 + \frac{\sin\theta}{2} \delta_2^\mathbf{A}
                                 + \frac{-\cos \theta + 1}{2} \delta_3^\mathbf{A}
                                 + \frac{\cos \theta + 1}{2} \delta_4^\mathbf{A}, \\
    \tilde\delta_5^\mathbf{A'} = \delta_5^\mathbf{A},~
    \tilde\delta_6^\mathbf{A'} =  \delta_6^\mathbf{A},~
    \tilde\delta_7^\mathbf{A'} = \delta_7^\mathbf{A} \cos (2\theta),~
    \tilde\delta_8^\mathbf{A'} = \delta_8^\mathbf{A} \cos (2\theta).
\end{gathered}
\end{math}
}
\end{equation}

\textbf{Proof:} 
Assume the bounding box $\mathbf{y}$ is centered at $\mathbf{c} \in \mathbb{R}^{3 \times 1}$ with dimension $\mathbf{d} \in \mathbb{R}^{3 \times 1}$ and yaw $\phi \in \mathbb{R}$. 
Since the spatial translation does not affect the relative position of voxels and bounding boxes, here we can only consider the effect of random rotation around the upright-axis $\theta$.
Since we have:
\begin{equation}
\begin{bmatrix}
\tilde\delta_1^\mathbf{A'}  \\
\tilde\delta_2^\mathbf{A'}  \\
\tilde\delta_3^\mathbf{A'}  \\
\tilde\delta_4^\mathbf{A'} 
\end{bmatrix} 
= 
\begin{bmatrix}
\tilde{x} - \tilde{\hat {x}}  \\
\tilde{\hat {x}} - \tilde{x}  \\
\tilde{y} - \tilde{\hat {y}}  \\
\tilde{\hat {y}} - \tilde{y}
\end{bmatrix}  
+
\begin{bmatrix}
\frac{1}{2}w  \\
\frac{1}{2}w  \\
\frac{1}{2}h  \\
\frac{1}{2}h 
\end{bmatrix} 
\\
=
\widetilde{\begin{bmatrix}
{x} - {\hat {x}}  \\
{\hat {x}} - {x}  \\
{y} - {\hat {y}}  \\
{\hat {y}} - {y}
\end{bmatrix} }
+
\begin{bmatrix}
\frac{1}{2}w  \\
\frac{1}{2}w  \\
\frac{1}{2}h  \\
\frac{1}{2}h 
\end{bmatrix} 
\end{equation}

By noting that,
\begin{equation}
\begin{aligned}
\widetilde{\begin{bmatrix}
{x} - {\hat {x}}  \\
{y} - {\hat {y}}
\end{bmatrix} }
&=
\begin{bmatrix}
 \cos \theta & -\sin \theta \\
 \sin \theta & \cos \theta
\end{bmatrix}
\begin{bmatrix}
{x} - {\hat {x}}   \\
{y} - {\hat {y}}
\end{bmatrix} \\
&=
\frac{1}{2}
\begin{bmatrix}
 \cos \theta & -\sin \theta \\
 \sin \theta & \cos \theta
\end{bmatrix}
\begin{bmatrix}
\delta_1 - \delta_2   \\
\delta_3 - \delta_4
\end{bmatrix} \\
& = \frac{1}{2}  
\begin{bmatrix}
 \cos \theta & -\cos \theta & -\sin \theta & \sin \theta \\
 \sin \theta & -\sin \theta &  \cos \theta & -\cos \theta
\end{bmatrix}
\begin{bmatrix}
\delta_1 \\ \delta_2   \\
\delta_3 \\ \delta_4
\end{bmatrix}
\end{aligned}
\end{equation}

And that,
\begin{align}
\begin{bmatrix}
w  \\
h
\end{bmatrix}
=
\begin{bmatrix}
\delta_1 + \delta_2   \\
\delta_3 + \delta_4
\end{bmatrix} = 
\begin{bmatrix}
 1 & 1  & 0  & 0 \\
 0 & 0 &  1 & 1
\end{bmatrix}
\begin{bmatrix}
\delta_1 \\ \delta_2   \\
\delta_3 \\ \delta_4
\end{bmatrix}
\end{align}
Then we have,
\begin{equation}
\resizebox{.9\linewidth}{!}{
\begin{math}
\begin{gathered}
    \tilde\delta_1^\mathbf{A'} = \frac{\cos \theta + 1}{2} \delta_1^\mathbf{A}
                                 + \frac{-\cos \theta + 1}{2} \delta_2^\mathbf{A}
                                 + \frac{-\sin\theta}{2} \delta_3^\mathbf{A}
                                 + \frac{\sin\theta}{2} \delta_4^\mathbf{A}, \\
    \tilde\delta_2^\mathbf{A'} = \frac{-\cos \theta + 1}{2} \delta_1^\mathbf{A}
                                 + \frac{\cos \theta + 1}{2} \delta_2^\mathbf{A}
                                 + \frac{\sin\theta}{2} \delta_3^\mathbf{A}
                                 + \frac{-\sin\theta}{2} \delta_4^\mathbf{A}, \\
    \tilde\delta_3^\mathbf{A'} = \frac{\sin\theta}{2} \delta_1^\mathbf{A} 
                                 + \frac{-\sin\theta}{2} \delta_2^\mathbf{A}
                                 + \frac{\cos \theta + 1}{2} \delta_3^\mathbf{A}
                                 + \frac{-\cos \theta + 1}{2} \delta_4^\mathbf{A}, \\
    \tilde\delta_4^\mathbf{A'} = \frac{-\sin\theta}{2} \delta_1^\mathbf{A} 
                                 + \frac{\sin\theta}{2} \delta_2^\mathbf{A}
                                 + \frac{-\cos \theta + 1}{2} \delta_3^\mathbf{A}
                                 + \frac{\cos \theta + 1}{2} \delta_4^\mathbf{A}.
\end{gathered}
\end{math}
}
\end{equation}
The rotation around the upright-axis does not affect $z$-coordinates, so it is trivial that,
\begin{align}
    \tilde\delta_5^\mathbf{A'} = \delta_5^\mathbf{A}, \tilde\delta_6^\mathbf{A'} =  \delta_6^\mathbf{A}.
\end{align}
The rotation does transform the yaw angle from $\phi$ to $\phi - \theta$, hence we have:
\begin{equation}
    \begin{aligned}
        \tilde\delta_7^\mathbf{A'} & = \log (\frac{w}{l}) \sin (2\phi - 2\theta) \\
                                &= \log (\frac{w}{l}) \left(\sin (2\phi) \cos (2\theta) - \sin (2\theta) \cos (2\phi)\right) \\
        \tilde\delta_8^\mathbf{A'} & = \log (\frac{w}{l}) \cos (2\phi - 2\theta) \\
                                &= \log (\frac{w}{l}) (\cos (2\phi) \cos (2\theta) + \sin (2\theta) \sin (2\phi))
    \end{aligned}
\end{equation}
By noting that when $\theta \in \{ \frac{k\pi}{2} \}_{k=0}^{3}$, $\sin(2\theta) \equiv 0$.
That produces,
\begin{equation}
    \begin{aligned}
        \tilde\delta_7^\mathbf{A'} = \log (\frac{w}{l}) \sin (2\phi) \cos (2\theta) = \delta_7^\mathbf{A} \cos (2\theta) \\
        \tilde\delta_8^\mathbf{A'} = \log (\frac{w}{l}) \cos (2\phi) \cos (2\theta) = \delta_8^\mathbf{A} \cos (2\theta) \\
    \end{aligned}
\end{equation}

Eq.~16, Eq.~17 and Eq.~19 can be combined to form Eq.~12. $\square$

\paragraph{Proof of Equation 8} 
Here we prove: 
\begin{equation}
    \label{eq:equal0}
    [\{\mathbf{A}\} \mathbf{R}_{\theta} + \{\Delta \mathbf{r}\} + \vec{\mathbf{r}'}] = \mathbf{0}
\end{equation}
We start from Eq.~7 from the main paper:
\begin{equation}
    \label{eq:eqv}
    [ \mathbf{A}\mathbf{R}_{\theta, \Delta \mathbf{r}}  + \vec{\mathbf{r}'} ] = [ [\mathbf{A}] \mathbf{R}_{\theta, \Delta \mathbf{r}}]
\end{equation}
By defactoring $\mathbf{A}\mathbf{R}_{\theta, \Delta \mathbf{r}}$ into $\mathbf{A}\mathbf{R}_{\theta} + \Delta \mathbf{r}$, we have:
\begin{equation}
    \label{eq:eqv}
    [ \mathbf{A}\mathbf{R}_{\theta} + \Delta \mathbf{r}  + \vec{\mathbf{r}'} ] = [ [\mathbf{A}] \mathbf{R}_{\theta} + \Delta \mathbf{r}]
\end{equation}
Noting all elements in $[\mathbf{A}]$ are integers, hence by assuming $\theta \in \{ \frac{k\pi}{2} \}_{k=0}^{3}$ and applying Lemma.~2, all elements in $[\mathbf{A}] \mathbf{R}_{\theta}$ are also integers.
Then by Lemma.~1, we have:
\begin{equation}
    \label{eq:eqv}
    [ \mathbf{A}\mathbf{R}_{\theta} + \Delta \mathbf{r}  + \vec{\mathbf{r}'} ] = [\mathbf{A}] \mathbf{R}_{\theta} + [  \Delta \mathbf{r}]
\end{equation}
Leveraging the property that $\mathbf{X} = [\mathbf{X}] + \{\mathbf{X}\}$, we have:
\begin{equation}
    \label{eq:eqv}
    [ ([\mathbf{A}] + \{\mathbf{A}\})\mathbf{R}_{\theta} + [\Delta \mathbf{r}] + \{\Delta \mathbf{r}\}  + \vec{\mathbf{r}'} ] = [\mathbf{A}] \mathbf{R}_{\theta} + [  \Delta \mathbf{r}]
\end{equation}
A simple deformation of this equation yields:
\begin{equation}
    \label{eq:eqv}
    [ [\mathbf{A}]\mathbf{R}_{\theta} + [\Delta \mathbf{r}] + \{\mathbf{A}\}\mathbf{R}_{\theta} + \{\Delta \mathbf{r}\}  + \vec{\mathbf{r}'} ] = [\mathbf{A}] \mathbf{R}_{\theta} + [  \Delta \mathbf{r}]
\end{equation}
By Lemma.~1, we move the term $[\mathbf{A}] \mathbf{R}_{\theta} + [  \Delta \mathbf{r}]$ out of the left-hand side, and that yields:
\begin{equation}
    \label{eq:equal0}
    [\{\mathbf{A}\} \mathbf{R}_{\theta} + \{\Delta \mathbf{r}\} + \vec{\mathbf{r}'}] = \mathbf{0}
\end{equation}
That is the exact form as Eq.~8 in the original paper.$\square$

\paragraph{Solution to Equation 10}
Here we find the solution $\boldsymbol\gamma_0$ of:
\begin{equation}
    \label{eq:final-eq}
    \boldsymbol\gamma_0 = \rm{argmin}_{\boldsymbol\gamma\in [0, S_\text{v}]^3} \ || \boldsymbol\gamma - \{\mathbf{A}\} \mathbf{R}_{\theta} - \{\Delta \mathbf{r}\}  ||_2
\end{equation}

We start by considering cases for unary functions. We find the solution $\boldsymbol\phi_0$ of:
\begin{equation}
    \label{eq:final-eq}
    \boldsymbol\phi_0 = \rm{argmin}_{\boldsymbol\phi\in [a, b]} \ || \boldsymbol\phi - M||_2
\end{equation}
The solution is straight-forward. It denotes the closest value in $[a, b]$ to a fixed value $M$. We represent the solution to this problem as:
\begin{align}
    \rm{clamp}(M, a, b) = \begin{cases}
        a,& M < a, \\
        M, & a \le M < b, \\
        b, & M \ge b.
    \end{cases}
\end{align}
Since in the target function of this problem, the three axes are uncorrelated, we can break this problem to a problem set of three problems each equivalent to Eq.~10.
We can extend the clamping function to a vector version, namely for any $0 \le i < \text{len}(\mathbf{M})$ :
\begin{align}
    \rm{clamp}(\mathbf{M}, \mathbf{a}, \mathbf{b})_i = \rm{clamp}(\mathbf{M}_i, \mathbf{a}_i, \mathbf{b}_i)
\end{align}
Then the closed-form solution of $\boldsymbol\gamma_0$ can be formulated as:
\begin{equation}
    \boldsymbol\gamma_0 = \rm{clamp}(\mathbf{M} = \{\mathbf{A}\} \mathbf{R}_{\theta} + \{\Delta \mathbf{r}\}, (0,0,0), (S_v , S_v, S_v))
\end{equation}

That is the solution to the original problem. $\square$

\section{Hyperparameter Study}

\begin{figure}[htbp]
    \centering
    \includegraphics[width=1.00\linewidth]{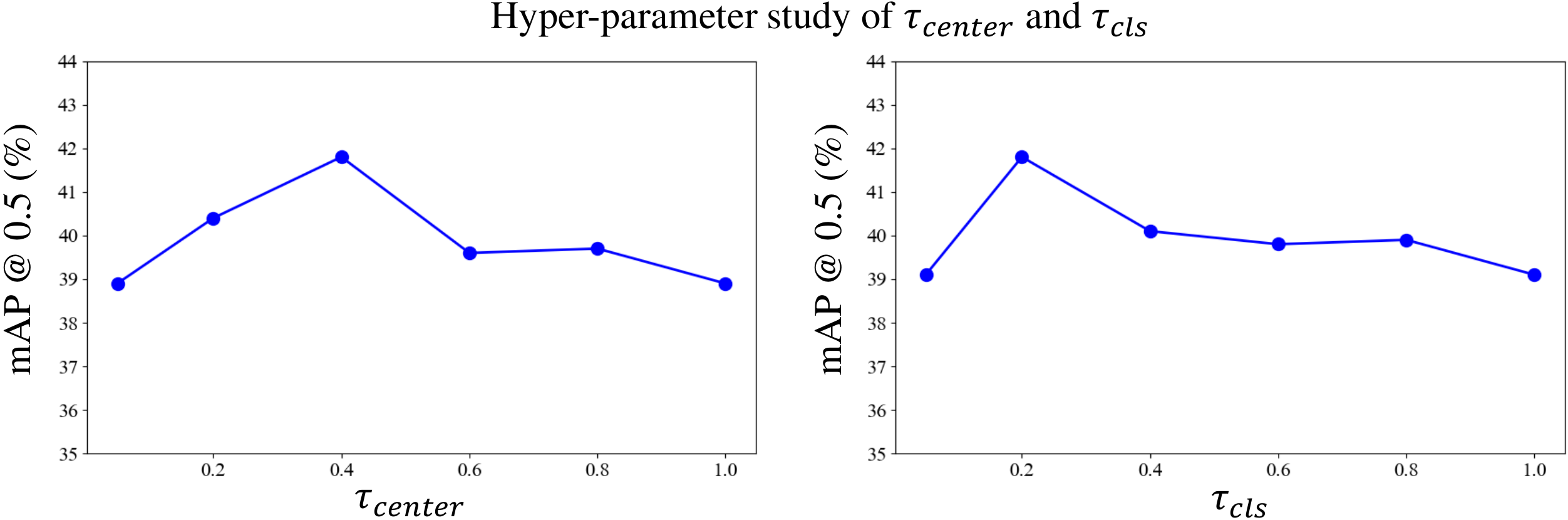}
    \caption{Hyper-parameter Study on $\tau_\text{center}$ and $\tau_\text{cls}$.
    }
    \label{fig:hyper}
\end{figure}

\textbf{$\tau_\text{center}$ and $\tau_\text{cls}$.} 
We conducted a hyperparameter study (Fig.~\ref{fig:hyper}) on $\tau_\text{center}$ and $\tau_\text{cls}$. These two hyperparameters are utilized to filter the initially matched set and provide matching pairs that offer less noisy supervision. 
Finding the optimal values involves a trade-off, as setting the values too low introduces noisy supervision, while setting them too high reduces the number of matched pairs.

\begin{table}[htbp]
\centering
\resizebox{0.8\columnwidth}{!}{
\begin{tblr}{
  cells = {c},
  cell{1-6}{1} = {l},
  hline{1,5,7}={},
  hline{2}={}
}
Backbone (Semi-supervised Setting) & $\rm mAP@0.25$ & $\rm mAP@0.50$ \\
FCAF3D (baseline)& 58.2 & 42.1   \\
FCAF3D (+ Sparse Proposal Matching) & 62.0 & 44.2  \\
FCAF3D (+ Dense Matching, \textit{ours DQS3D}) & \textbf{64.3} & \textbf{48.5}   \\
TR3D (baseline) & 62.5 & 46.8   \\
TR3D (+ Dense Matching) & \textbf{65.4} & \textbf{49.9 }
\end{tblr}
}
\caption{Comparison of \textit{Dense Matching} and \textit{Proposal Matching} Strategies with \textit{Different Backbones} on ScanNet Dataset (20\% Labeled). \textit{Proposal matching} involves filtering teacher proposals and matching them with the nearest-center student predictions, while dense matching establishes matching based on spatially-aligned voxel anchors and then applies filtering. In dense matching, the proposed \textbf{Q}uantization \textbf{E}rror \textbf{C}orrection module is enabled.}
\label{tab:experiment}
\end{table}

\textbf{Different Backbones.}
We conducted experiments (Tab.~\ref{tab:experiment}) that show the superiority of dense matching over proposal matching. 
We argue that the success is originate from addressing issues like\textit{ \textcolor[rgb]{0.4375 0.1875 0.625}{no supervision}} and \textit{\textcolor[rgb]{0.9257, 0.5, 0.1914}{multiple supervision}} problems, which we also qualitatively illustrate in Fig.~\ref{fig:viz_result}.
Note that dense matching is applicable only to recent SOTA voxel-based detectors, not common two-stage proposal-based detectors based on Transformer or heatmaps. 
Hence we used TR3D (Rukhovich et al.), with the hyperparameters reported in our manuscript without further tuning. 
Remarkably, we observed an improvement of +3.1\% on $\rm{mAP@0.50}$.

\section{Further Discussion}
\paragraph{Computational Complexity Analysis.} 
We utilized the NVIDIA GeForce RTX 2080Ti.
Training employed 4 GPUs (2 labeled and 2 unlabeled scenes per GPU card, occupying approximately 7.5GB per GPU) and took around 7 hours to converge.
In terms of inference speed, our system achieves 10.3 scenes per second on a single 2080Ti.

\paragraph{Limitation Analysis.}
The trade-off between memory and voxel size hampers our 3D detectors' performance in outdoor scenes, which is a common limitation in the family of sparse convolutional detectors.

\end{document}